%% file: main.tex
\documentclass[11pt]{article}

\usepackage[letterpaper,margin=1in]{geometry}
\usepackage[T1]{fontenc}
\usepackage[utf8]{inputenc}

\usepackage{amsmath}
\usepackage{amssymb}
\usepackage{amsthm}
\usepackage{mathtools}

\usepackage{newtxtext}
\usepackage{newtxmath}
\usepackage{bm}
\usepackage{bbm}

\usepackage{graphicx}
\usepackage[shortlabels]{enumitem}
\usepackage{microtype}
\usepackage{authblk}   

\allowdisplaybreaks

\usepackage[numbers,sort&compress]{natbib}
\bibpunct[, ]{[}{]}{,}{n}{}{,}%
\usepackage{xcolor}
\definecolor{linkcolour}{rgb}{0.05,0.25,0.55}
\usepackage[%
  colorlinks=true,
  linkcolor=linkcolour,
  citecolor=linkcolour,
  urlcolor=linkcolour,
  pdftitle={Optimal Convergence Rate for RVI Q-Learning},
  pdfkeywords={average-reward Markov decision processes, relative value iteration Q-learning, stochastic approximation, error bounds, finite-time bounds},
  pdfauthor={Avik Kar, Siddharth Chandak, Rahul Singh, Shalabh Bhatnagar}%
]{hyperref}

\theoremstyle{plain}
\newtheorem{theorem}{Theorem}
\newtheorem{lemma}{Lemma}
\newtheorem{proposition}{Proposition}
\newtheorem{corollary}{Corollary}
\newtheorem{assumption}{Assumption}

\theoremstyle{definition}
\newtheorem{example}{Example}
\newtheorem{definition}{Definition}
\newtheorem{remark}{Remark}

\newcommand{\eps}{\epsilon}
\newcommand{\ust}{^\star}

\newcommand{\bE}{\mathbb{E}}
\newcommand{\bP}{\mathbb{P}}
\newcommand{\bR}{\mathbb{R}}
\newcommand{\bo}{\mathbf{1}}

\newcommand{\cA}{\mathcal{A}}
\newcommand{\cE}{\mathcal{E}}
\newcommand{\cF}{\mathcal{F}}

\newcommand{\cS}{\mathcal{S}}
\newcommand{\cX}{\mathcal{X}}

\newcommand{\ol}[1]{\overline{#1}}
\newcommand{\flbr}[1]{\left\{#1\right\}}
\newcommand{\sqbr}[1]{\left[#1\right]}
\newcommand{\br}[1]{\left(#1\right)}
\newcommand{\abs}[1]{\left|#1\right|}
\newcommand{\norm}[1]{\left\Vert #1 \right\Vert}

\newcommand{\fix}{\operatorname{Fix}}

\title{\bfseries $O(1/n)$ Convergence Rate for RVI $Q$-Learning}

\author[1]{Avik Kar}
\author[2]{Siddharth Chandak}
\author[3]{Rahul Singh}
\author[1]{Shalabh Bhatnagar}

\affil[1]{Department of Computer Science and Automation, Indian Institute of Science, Bengaluru, India. \texttt{avikkar@iisc.ac.in}, \texttt{shalabh@iisc.ac.in}}
\affil[2]{Department of Electrical Engineering, Stanford University, Stanford, California, USA. \texttt{chandaks@stanford.edu}}
\affil[3]{EPITA, Laboratoire Recherche de l'EPITA, Paris, France. \texttt{rahulsingh0188@gmail.com}}

\date{}

\begin{document}

\maketitle

\begin{abstract}
    We establish finite-time convergence guarantee for synchronous relative value iteration (RVI) $Q$-learning for finite weakly communicating Markov decision processes (MDPs). For stochastic Krasnosel'ski\u{\i}--Mann iterations driven by a non-expansive, linearly regular map on a finite-dimensional normed space, we establish last-iterate mean-squared residual bounds under martingale-difference noise whose conditional second moment grows at most linearly with the squared residual. Suitable polynomial stepsizes of order $n^{-a}$ yield an $O(n^{-a})$ rate for $0<a\leq1$. 
    We verify linear regularity and the noise condition for the centered Bellman operator for finite weakly communicating MDPs. Applying the general result gives an $O(n^{-1})$ bound on the mean-squared Bellman span residual of RVI $Q$-learning iterates when a suitable stepsize sequence is used. The positive gap between optimal and suboptimal stationary deterministic policies then yields an $O(n^{-1})$ bound on the expected suboptimality of the greedy policy. Consequently, $O(\eps^{-1})$ synchronous iterations suffice for an $\eps$-optimal policy in expectation. These guarantees are instance-dependent, with constants that depend on the hitting-time parameters and the suboptimality gap of the second-best policy.
\end{abstract}


\smallskip

\input{tex_files/intro}
\input{tex_files/rvi}

\input{tex_files/km}
\input{tex_files/rvi_rates}
\input{tex_files/conclusion}

\appendix
\input{appendix/lin_regularity}
\input{appendix/mds_props}

\input{appendix/stepsize}


\bibliographystyle{plainnat}
\bibliography{refs}

\end{document}

%% file: tex_files/intro.tex
\section{Introduction}
\label{sec:intro}

Average-reward reinforcement learning (RL), in which the learning agent aims to maximize the long-run average reward, provides a natural model for sequential decision-making problems in systems where steady-state performance is crucial, and there is no prescribed terminal time, such as queueing networks, inventory control, robotics, power-grid management, and communication networks~\citep{puterman2014markov}. In contrast, discounted RL assigns geometrically decaying weights to future rewards and therefore does not directly capture the long-run steady-state performance of a policy. Discounting, however, offers a key analytical advantage. The value function is the unique fixed point of the Bellman operator, which is a contraction~\citep[Chapter 6]{puterman2014markov}. Similar properties hold in undiscounted problems when the state process reaches a terminal state almost surely under every policy~\citep{bertsekas1991analysis}. This contractive structure underlies much of the finite-time theory for value-based RL algorithms such as \(Q\)-learning and TD-learning~\citep{chandak2022concentration,wainwright2019stochastic,chen2020finite,srikant2019finite,li2024minimax}. Average-reward RL is inherently more challenging from this perspective. In the absence of discounting, the Bellman operator and related maps, such as the RVI map, are generally only non-expansive. Moreover, solutions to the average-reward optimality equation~\eqref{eq:aroe}, i.e., the fixed points of the centered Bellman operator~\eqref{def:aux_map}, are never unique. For unichain MDPs, the solution set has one degree of freedom~\citep[Chapter 8]{puterman2014markov}, whereas for the weakly communicating MDPs studied here, the solution set is, in general, higher dimensional~\citep{schweitzer1978functional,wan2024convergence}.

RVI $Q$-learning, introduced by \citet{abounadi2001learning}, is one of the earliest provably convergent model-free value-based algorithms for average-reward RL. It updates the estimate of the optimal $Q$-function with a sampled Bellman step after subtracting an iterate-dependent normalization term from it~\eqref{eq:rvi_update}. The normalization term removes the undetermined additive component of the $Q$-function and stabilizes the iterates. Under certain regularity conditions on the normalization functional and standard choices of the stepsize sequence, \citet{abounadi2001learning} established almost-sure convergence of both synchronous and asynchronous RVI $Q$-iterates for unichain MDPs with a common state recurrent under every stationary policy. Their result is asymptotic and does not quantify the number of iterations required for obtaining a $Q$-iterate that nearly solves the average reward optimality equation~\eqref{eq:aroe}, or for obtaining a nearly optimal policy. The goal of the current work is to provide such a finite-time characterization for synchronous RVI $Q$-learning by studying the fixed-point residual, which upper bounds the suboptimality of the greedy policy induced by the current $Q$-iterate.

Due to the non-expansive nature of the relevant operators, the finite-time analysis available for contractive settings cannot be applied here. Although the average-reward Bellman operator is non-expansive under the sup-norm, the RVI map is non-expansive only under the span seminorm. A standard approach is to analyze an auxiliary iteration driven by the centered Bellman operator, in which the constant optimal average reward replaces the iterate-dependent normalization term in RVI $Q$-learning. The operator is non-expansive under the sup-norm, and the iterates, when run with the same samples, have the same Bellman span residuals as the RVI iterates. Hence, obtaining finite-time guarantees for RVI $Q$-learning reduces to analyzing stochastic Krasnosel'ski\u{\i}--Mann (KM) iterations with non-expansive operators under an arbitrary norm.

The following two lines of work address finite-time guarantees for average-reward $Q$-learning. Analyses via the theory of stochastic KM iterations assume nothing beyond non-expansiveness.~\citet{bravo2024stochastic} obtain an expected residual of order $\widetilde{O}(n^{-1/6})$ after $n$ iterations for uniformly bounded noise variance, where $\widetilde O$ hides logarithmic factors. This rate degrades to $O(n^{-1/10+\eps})$, for any $\eps \in (0,1/10)$, once the iterate-dependence of the RVI sampling noise is accounted for. On the other hand, a body of work assumes a contraction property and obtains the faster rate $O(n^{-1/2})$.~\citet{chen2025seminorm} assume a seminorm contraction and obtain $O(n^{-1/2})$; \citet{chen2025adaptive} obtain last-iterate guarantees under a span-contraction assumption with adaptive stepsizes; and \citet{chen2026contraction} construct an instance-dependent seminorm after a lazy transformation of the dynamics and obtain an $\widetilde{O}(\eps^{-2})$ sample complexity under a reachability condition.

Instead of exploiting only the non-expansive property of the RVI $Q$-learning operator, or imposing any contractive structure, the current work shows that a third property, called \emph{linear regularity}~\citep{bauschke2015linear}, holds for the centered Bellman operator if the MDP is weakly communicating.

\begin{definition}[Linearly regular operator]\label{def:linear_regular}
    Let $(\cX ,\norm{\cdot})$ be a normed space, and let $T: \cX \to \cX$ be an operator with a non-empty set of fixed points $\fix(T) := \{x \in \cX : T x = x\}$. $T$ is linearly regular if there exists a constant $\kappa > 0$ such that
    \begin{align*}
        \inf_{x\ust \in \fix(T)}\norm{x - x\ust} \leq \kappa\norm{x - Tx}, ~\forall x \in \cX.
    \end{align*}
\end{definition}

Further, we show that the conditional second moment of the RVI sampling noise is at most a constant times one plus the squared residual of the current iterate. Using these two properties of the operator and the noise, the current work shows that the mean-squared residuals of the stochastic KM iterates decay at a rate of $O(n^{-1})$. The same rate holds for the expected suboptimality of the policy greedy with respect to the RVI $Q$-iterates.

\paragraph{Contributions.} Our main results are summarized in the following three points.
\begin{enumerate}[(i)]
    \item For the stochastic KM iteration $x_n=(1-\alpha_n)x_{n-1}+\alpha_n(Tx_{n-1}+M_n)$ driven by a non-expansive, linearly regular map $T$ with linear regularity constant $\kappa$ on an arbitrary finite-dimensional normed space, with stepsizes $\alpha_n$ and martingale-difference noise $M_n$ whose conditional second moment is bounded by $\sigma^2(1+\norm{x_{n-1}-Tx_{n-1}}^2)$, we prove a last-iterate bound on the mean-squared residual (Theorem~\ref{thm:km_main}). Polynomial stepsizes $c(n+K_0)^{-a}$ with a fixed $a \in (0,1]$ attain the rate of $O(n^{-a})$ .
    \item We prove that the centered Bellman operator $\bar H$~\eqref{def:aux_map} for every finite weakly communicating MDP is linearly regular with the explicit constant $K$, defined in \eqref{def:K} (Proposition~\ref{prop:rvi_error_bound}).~We also prove that the conditional second moment of its sampling noise is bounded by a constant times one plus the squared residual (Proposition~\ref{prop:rvi_noise}).
    \item Combining (i) and (ii), we obtain an $O(n^{-1})$ bound on the mean-squared Bellman span residual of the RVI $Q$-iterates, with a prefactor of order $K^6(1+D)^2\log(2|\cS||\cA|)$ (Theorem~\ref{thm:rvi_main}). The residual-to-policy relation~\eqref{eq:greedy_loss} gives the same rate for expected suboptimality of the greedy policy, with the additional factor $1/\Delta_{\rm pol}$ in its prefactor. Thus, $O(\eps^{-1})$ synchronous iterations suffice for the greedy policy to be $\eps$-optimal in expectation (Corollary~\ref{cor:sample_complexity}).
\end{enumerate}

\paragraph{Overview of the argument.} To prove (i), we first consider the iteration without sampling noise. Because of linear regularity, over a sufficiently large number of iterations, called a block, the distance to the nearest fixed point decreases by a fraction. We then compare the noisy and noiseless iterations starting from the same point. The comparison shows a similar decrease in the mean-squared distance, with an additional error proportional to the step size at the start of the block. In the next block, the accumulated error is reduced again, while new noise contributes another error term. The stepsize conditions ensure that this repeated reduction suppresses the errors introduced in earlier blocks. Consequently, the mean-squared distance, and hence the mean-squared residual, is bounded by a constant times the current stepsize plus a decaying contribution from the initial distance.

For (ii), we first obtain a preliminary bound on the distance to the solution set in terms of the residual, along with an additive constant. Whenever the residual is above a strictly positive threshold, the additive constant can be absorbed into the residual term at the expense of a larger constant. We set the threshold to be the \emph{policy gap}, which is the smallest non-zero loss in average reward among stationary deterministic policies. Wherever the residual is smaller than the policy gap, the corresponding greedy policy must be optimal. We use this property to construct a nearby solution. Given a $Q$-function, let the value of a state be the largest $Q$-value over actions at that state. We show that these values can be raised by an amount proportional to the residual so that subsequent Bellman updates, with the optimal average reward subtracted, only decrease them. The updated values remain bounded below, so they converge to an exact state-value solution. To bound its distance from the original values, we use the closed groups of states in which the greedy policy eventually remains, called its recurrent classes. On these classes, optimality forces the Bellman updates to leave the raised values unchanged. For other states, summing Bellman errors until one of these classes is reached bounds how far the limiting values lie below the original values. The expected hitting times make this bound proportional to the residual. Together with the bound on the initial upward adjustment, this shows that the limiting state values differ from the original values by an amount proportional to the residual. Converting this state-value solution back to a $Q$-function gives the nearby exact solution needed for linear regularity. Separately, we bound the spread of state values across states in terms of the residual. This controls the sampling noise, which comes from the bounded reward and the value of the sampled next state.

For (iii), we run the auxiliary and RVI iterations using the same sampled transitions. Their iterates differ only by a constant added to every entry, which leaves the Bellman span residual unchanged. The bound from (i) therefore gives a mean-squared residual bound for RVI. For each sample outcome, the greedy policy's loss is at most the Bellman span residual and is either zero or at least the policy gap. These two facts bound the loss by the squared residual divided by the gap. Taking expectations gives the claimed $O(n^{-1})$ rate for the stated stepsizes of order $1/n$, with constants depending on the fixed MDP.

\paragraph{Related work.} \citet{abounadi2001learning} establish convergence of RVI $Q$-learning for unichain MDPs with a common state recurrent under every stationary policy, and \citet{wan2024convergence} extend this result to weakly communicating MDPs. \citet{wan2021learning} introduce differential $Q$-learning, which avoids a normalization functional, and prove its asymptotic convergence. For finite-time guarantees, \citet{zhang2021finite} analyze tabular average-reward $Q$-learning under span-seminorm contraction. \citet{chen2025seminorm} develop a seminorm Lyapunov framework yielding an $O(n^{-1/2})$ expected span-seminorm error bound. \citet{chen2025adaptive} obtain asynchronous last-iterate guarantees with adaptive stepsizes under span contraction. \citet{chen2026contraction} construct an instance-dependent seminorm after a lazy transformation and obtain $\widetilde{O}(\eps^{-2})$ sample complexity under a reachability condition. For weakly communicating MDPs with a generative model, \citet{zurek2024span} use a discounted reduction to obtain the minimax-optimal sample complexity $\widetilde{O}(|\cS||\cA|\mathsf{H}/\eps^2)$, where $\mathsf{H}$ is the span of the optimal value function. Our $O(\eps^{-1})$ iteration bound has instance-dependent constants, so it does not imply an improvement over this minimax benchmark.

\citet{bravo2024stochastic} analyze stochastic KM iterations on arbitrary normed spaces, obtaining an expected residual rate of $\widetilde{O}(n^{-1/6})$ under uniformly bounded noise variance. For RVI $Q$-learning, iterate-dependent noise yields the slower rate $O(n^{-1/10+\eta})$ for any $\eta\in(0,1/10)$. \citet{blaser2026nonexpansive} study nonexpansive stochastic approximation with Markovian noise and apply their results to average-reward TD learning. Our analysis instead exploits linear regularity and residual-dependent martingale noise to obtain an $O(n^{-1})$ bound on the mean-squared residual.

Alternative algorithms include the adaptive-discount $Q$-learning method of \citet{jin2024feasible}, which provides finite-sample guarantees without prior knowledge of problem parameters. \citet{lee2025near} combine Halpern's anchored iteration with recursive sampling to obtain an $\eps$-optimal policy using $\widetilde{O}(|\cS||\cA|\mathsf{H}^2/\eps^2)$ samples without requiring prior knowledge of $\mathsf{H}$.

\paragraph{Organization.} Section~\ref{sec:rvi} introduces the problem and algorithm and states the main results. Section~\ref{sec:km} develops a general convergence bound for stochastic KM iterations. Section~\ref{sec:rvi_rates} applies this bound to RVI $Q$-learning, establishes the required linear regularity and noise growth condition, and proves the convergence and sample complexity guarantees. Section~\ref{sec:conclusion} concludes. Appendices~\ref{sec:app_supporting}--\ref{sec:app_aux} contain supporting lemmas, a martingale inequality, and stepsize estimates, respectively.

%% file: tex_files/rvi.tex
\section{Setup and Main Results}
\label{sec:rvi}

\subsection{Average-Reward MDPs}
\label{subsec:amdp}

\paragraph{Notation.} For a finite dimensional vector $x$, $\norm{x}_{\rm sp} := \max_i x(i) - \min_i x(i)$ denotes the span seminorm. For a finite set $\cX$, $\Delta_\cX$ is the probability simplex on $\cX$.

Let $(\cS,\cA,p,r)$ be an MDP, where $\cS$ and $\cA$ are finite state and action spaces, respectively, $p(s,a,\cdot) \in \Delta_\cS$ governs the state transitions and $r: \cS \times \cA \times \cS \to [0,1]$ is the reward function.~For a stationary policy $\pi: \cS \to \Delta_\cA$, we write $\pi(a|s)$ for the probability that $\pi$ chooses action $a$ in state $s$. Its average reward starting from state $s$ is
\begin{align*}
    J(\pi,s):=\liminf_{n\to\infty}\frac{1}{n}\bE^\pi_s\sqbr{\sum_{k=0}^{n-1}r(S_k,A_k,S_{k+1})},
\end{align*}
where $S_k$ and $A_k$ denote the state and the action at time $k$, and $\bE^\pi_s$ denotes expectation under the conditions that $S_0 = s$ and $A_k \sim \pi(\cdot|S_k)$ for every $k\geq0$. The MDP satisfies the following assumption.

\begin{assumption}[Weakly communicating]\label{assum:wc}
    The state space admits a partition $\cS = \cS_{\rm c} \cup \cS_{\rm t}$, where $\cS_{\rm c}$ is closed, i.e., $\sum_{s'\in\cS_{\rm c}}{p(s,a,s')} = 1$ for every $s \in \cS_{\rm c}$ and $a \in \cA$, every state of $\cS_{\rm c}$ is accessible from every other state of $\cS_{\rm c}$ under some stationary deterministic policy, and every state of the possibly empty set $\cS_{\rm t}$ is transient under every stationary policy.
\end{assumption}

The class of weakly communicating MDPs contains the families of ergodic, unichain, and communicating MDPs; see \citet[Section 8.3]{puterman2014markov}.~Under the above assumption, the optimal average reward $\sup_\pi{J(\pi,s)}$ is independent of the initial state $s$ and we denote the scalar value by $J\ust$.~The average-reward optimality equation~(AROE) is
\begin{align}
    J + Q(s,a) = \sum_{s' \in \cS}{p(s,a,s') \br{r(s,a,s') + V_Q(s')}},\quad \forall (s,a), \label{eq:aroe}
\end{align}
where $V_Q(s) := \max_{a\in \cA}{Q(s, a)}$. Assumption~\ref{assum:wc} is sufficient for the existence of $Q\ust \in \bR^{\cS \times \cA}$ such that $(J\ust, Q\ust)$ satisfies the AROE. In that case, $(J\ust, Q\ust + c \bo)$ also satisfies the AROE for every $c \in \bR$, where $\bo$ denotes the $\cS \times \cA$-dimensional all-ones vector.~These shifts need not exhaust the solution set. The AROE of a weakly communicating MDP may admit solutions that do not differ by a constant vector~\citep{schweitzer1978functional,wan2024convergence}. A policy that chooses an action from $\arg\max_a Q\ust(s,a)$ in each state $s$ is optimal. See \citet{puterman2014markov} for more details on average reward MDPs.

\paragraph{Useful MDP parameters.} Let $\Pi_{\mathrm{SD}}$ be the set of stationary deterministic policies, and let $\tau_A := \inf\{k \geq 0: S_k \in A\}$ and $\tau_{s'} := \tau_{\{s'\}}$. Define the following complexity measure of weakly communicating MDPs,
\begin{align}
    D := \max\flbr{\max_{s\neq s' \in \cS_{\rm c}}{\min_{\pi\in\Pi_{\mathrm{SD}}}{\bE_s^\pi[\tau_{s'}]}},~ \max_{\pi\in\Pi_{\mathrm{SD}}}{\max_{s\in\cS_{\rm t}}{\bE_s^\pi[\tau_{\cS_{\rm c}}]}}}, \label{def:D}
\end{align}
with the convention that the maximum of an empty set is $0$.~The first term is the diameter of the closed communicating set, and the second is the worst-case expected time to reach the closed communicating set from any transient state. The policy gap $\Delta_{\rm pol}$ is the smallest non-zero sub-optimality of stationary deterministic policies,
\begin{align}
    \Delta_{\rm pol}:=\min_{\pi\in\Pi_{\mathrm{SD}}:\ J\ust > J(\pi,s)\text{ for some }s}\ \max_{s'\in\cS}\br{J\ust-J(\pi,s')},
    \label{def:policy_gap}
\end{align}
with $\Delta_{\rm pol}:=1$ if every $\pi\in\Pi_{\mathrm{SD}}$ is optimal from every state. Since the policy set is finite and rewards lie in $[0,1]$, we have $0<\Delta_{\rm pol} \leq 1$. Thus, a stationary deterministic policy whose sub-optimality is less than $\Delta_{\rm pol}$ from every state is optimal from every state. Let $\bP^\pi_s$ denote the probability conditioned by $S_0 = s$ and $A_k \sim \pi(\cdot|S_k)$ for every $k \geq 0$. Define
\begin{align}
    \mathcal T:=
    \max_{\substack{\pi\in\Pi_{\mathrm{SD}},\ s\in\cS,\ \varnothing\ne C\subseteq\cS\\
    J(\pi,s')=J\ust\ \text{for every }s'\in\cS\\
    \bP_s^\pi(\tau_C<\infty)=1}} \bE_s^\pi[\tau_C].
    \label{def:finite_hitting_time}
\end{align}
Thus, $\mathcal T$ is the largest expected hitting time among stationary deterministic policies that are optimal from every state, initial states, and target sets for which the target is reached almost surely. This maximum is finite. Indeed, if $\bP_s^\pi(\tau_C<\infty)=1$, then every state reachable from $s$ before hitting $C$ has a path to $C$ of at most $|\cS|$ transitions with positive probability under policy $\pi$. Hence, conditionally on not yet hitting $C$, the probability of reaching $C$ within the next $|\cS|$ steps is bounded below by a positive number, so $\tau_C$ has a geometric tail and a finite expectation. $\mathcal{T}$ is finite because there are only finitely many triplets $(\pi,s,C)$.


\subsection{Synchronous RVI \texorpdfstring{$Q$}{Q}-Learning}
\label{subsec:rvi_algo}

\citet{abounadi2001learning} introduced the synchronous and asynchronous RVI $Q$-learning algorithms for learning a solution to the AROE. Synchronous RVI $Q$-learning updates the estimate of $Q\ust$ as follows: for all $(s,a) \in \cS \times \cA$,
\begin{align}
    Q_{n+1}(s,a) = Q_n(s,a) + \alpha_{n+1}\br{r(s,a,Y_{s,a}^{n+1})+V_{Q_n}(Y_{s,a}^{n+1}) - Q_n(s,a) - f(Q_n)}, \label{eq:rvi_update}
\end{align}
where, for each $n\geq0$, the random variables $\{Y^{n+1}_{s,a}\}_{(s,a)}$ are independent, with $Y^{n+1}_{s,a} \sim p(s,a,\cdot)$, and are independent of $Y^{m}_{s,a}$, $m\leq n$, $(s,a)\in\mathcal{S}\times\mathcal{A}$. $\cF_n := \sigma(Q_0,Y^m_{s,a}:1\leq m\leq n,\ (s,a)\in\cS\times\cA)$ denotes the natural filtration associated with the iterates. $\{\alpha_n\}_{n\geq 1}$ denotes the stepsize sequence, and  $f:\bR^{\cS\times \cA}\to\bR$ is a continuous normalization functional.~A few candidate $f$'s are $f(Q) = \min_{s,a}Q(s,a)$, $f(Q) = \max_{s,a}Q(s,a)$, $f(Q) = Q(s_0,a_0)$ for some fixed $(s_0, a_0) \in \cS \times \cA$, and $f(Q) = \frac{1}{|\cS||\cA|}\sum_{s,a}Q(s,a)$.

\citet{abounadi2001learning} showed that for unichain MDPs, $(f(Q_n),Q_n)$, $n\geq 0$, almost surely converges to a unique solution of \eqref{eq:aroe}, $(J\ust,Q\ust)$ such that $f(Q\ust) = J\ust$ if the functional $f$ satisfies certain regularity conditions and if the stepsizes satisfy
\begin{align*}
    \sum_{k\geq 1}{\alpha_k} = \infty, \qquad \sum_{k\geq 1}{\alpha^2_k}< \infty.
\end{align*}

Define the Bellman operator $B: \bR^{\cS\times \cA} \to \bR^{\cS\times \cA}$ as
\begin{align}
    (B Q)(s,a) := \sum_{s'\in \cS}{p(s,a,s') \br{r(s,a,s') + V_Q(s')}}, \label{def:bellman}
\end{align}
write $\ol r(s,a):=\sum_{s'\in\cS}p(s,a,s')r(s,a,s')$ for the expected one-step reward, and define the RVI map $H$ such that $H Q := B Q - f(Q) \bo$. The synchronous RVI $Q$-learning update~\eqref{eq:rvi_update} can alternatively be written as
\begin{align*}
    Q_{n+1}=Q_n + \alpha_{n+1} \br{H Q_n - Q_n + M_{n+1}(Q_n)},
\end{align*}
where
\begin{align*}
    M_{n+1}(Q) := \widehat{B}_{n+1} Q - B Q, \quad\mbox{with}\quad  (\widehat{B}_{n+1} Q)(s,a) := r(s,a,Y_{s,a}^{n+1})+V_{Q}(Y_{s,a}^{n+1}).
\end{align*}
Note that a solution $Q\ust$ of the AROE with $f(Q\ust) = J\ust$ is a fixed point of $H$.

The goal is to learn an optimal or near-optimal policy. Fix a deterministic tie-breaking rule, and let $\pi_Q$ be the resulting policy greedy with respect to $Q$. We measure performance by the suboptimality of the greedy policy,
\begin{align*}
    \Delta_n:=\max_{s\in\cS}\flbr{J\ust-J(\pi_{Q_n},s)}.
\end{align*}
For every $Q\in\bR^{\cS\times\cA}$, $0 \leq \max_{s\in\cS}\flbr{J\ust-J(\pi_Q,s)} \leq \norm{Q-BQ}_{\rm sp}$. See \citet[Proposition 2.1]{lee2025near} for a proof. Moreover, $HQ-BQ=-f(Q)\bo$ is a constant vector, so $\norm{Q-BQ}_{\rm sp}=\norm{Q-HQ}_{\rm sp}$. Thus, $\Delta_n\leq\norm{Q_n-HQ_n}_{\rm sp}$. However, since either $\Delta_n \geq \Delta_{\rm pol}$ or $\Delta_n = 0$, $\Delta_n \leq \Delta_n^2/\Delta_{\rm pol}$. Thus, 
\begin{align}
    \Delta_n\leq\frac{1}{\Delta_{\rm pol}}\norm{Q_n-HQ_n}^2_{\rm sp}. \label{eq:greedy_loss}
\end{align}
We bound the latter among the above two upperbounds of $\Delta_n$ in order to bound $\Delta_n$.

\subsection{Convergence Rates and Sample Complexity}
\label{subsec:rvi_results}

We now state the convergence rates and the sample complexity of synchronous RVI $Q$-learning. Recall $D$, $\Delta_{\rm pol}$ and $\mathcal{T}$ from \eqref{def:D}-\eqref{def:finite_hitting_time}. Set
\begin{align}
    K:=\max\flbr{1+\mathcal T,\ (1+3D)\br{1+\frac{1}{\Delta_{\rm pol}}}}.
    \label{def:K}
\end{align}

\begin{theorem}\label{thm:rvi_main}
    Let Assumption~\ref{assum:wc} hold, let $|\cS||\cA|\geq2$, and let $\{Q_n\}$ be generated by \eqref{eq:rvi_update} from a fixed $Q_0$ with the sampling model of Section~\ref{subsec:rvi_algo} and stepsizes $\alpha_n=c(n+K_0)^{-a}$, where $0<a\leq1$, $c>0$, and the integer $K_0\geq1$ satisfies
    \begin{align}
        cK_0^{1-a} \geq 2^8K^2, \qquad \alpha_1 \leq\frac{1}{\max\flbr{2,2^{6} C_1}},
        \label{eq:rvi_stepsize_cond}
    \end{align}
    where $C_1 := 2^{8}e(1+3D)^2(2^6K^2+1)^2\log(2|\cS||\cA|)$. Then, for every $n\geq0$,
    \begin{align*}
        \bE\norm{Q_n - H Q_n}^2_{\rm sp} \leq 100\cE(Q_0)^2 \exp\br{-\frac{\log(4)}{2^7K^2+\frac12}\sum_{k=1}^n\alpha_k} + \frac{45 C_1 c}{(n+K_0+1)^a},
    \end{align*}
    where $\cE(Q_0) = \inf_{q\in\fix(\bar H)}\norm{Q_0-q}_\infty$. By \eqref{eq:greedy_loss}, for every $n\geq0$,
    \begin{align}
        \bE\sqbr{\Delta_n}
        \leq \frac{100\cE(Q_0)^2}{\Delta_{\rm pol}}\exp\br{-\frac{\log(4)}{2^7K^2+\frac12}\sum_{k=1}^n\alpha_k}
        +\frac{45 C_1 c}{\Delta_{\rm pol}(n+K_0+1)^a}.
        \label{eq:rvi_main}
    \end{align}
    In particular:
    \begin{enumerate}[(i)]
        \item If $a\in(0,1)$, then \eqref{eq:rvi_stepsize_cond} holds for every sufficiently large $K_0$, and $\bE\sqbr{\Delta_n}\leq Cn^{-a}$ for every $n\geq1$, where $C$ depends only on the MDP, $Q_0$, $a$, $c$, and $K_0$.
        \item If $a=1$ and $c=2^8K^2$, then, with $p:=2^8K^2\log(4)/(2^7K^2+1/2)>1$,
        \begin{align}
            \bE\sqbr{\Delta_n}\leq \frac{100\cE(Q_0)^2}{\Delta_{\rm pol}}\br{\frac{K_0+1}{n+K_0+1}}^p
            +\frac{45\cdot2^{16} e K^2(2^6K^2+1)^2(1+3D)^2\log(2|\cS||\cA|)}{\Delta_{\rm pol}(n+K_0+1)}.
            \label{eq:rvi_hit_harmonic}
        \end{align}
    \end{enumerate}
\end{theorem}

Theorem~\ref{thm:rvi_main} improves on the $O(n^{-1/10+\eta})$ expected residual rate obtained for this algorithm by applying the iterate-dependent noise bound of \citet{bravo2024stochastic}, for any $\eta\in(0,1/10)$. For $a=1$, we obtain a convergence rate of $O(n^{-1})$ for each fixed MDP, with a prefactor of order $K^6(1+D)^2\log(2|\cS||\cA|)/\Delta_{\rm pol}$. The constants $K$ and $\Delta_{\rm pol}$ make the dependence on the MDP explicit. The proof is given in Section~\ref{sec:rvi_rates}. The following corollary converts the rate into a sample complexity result.

\begin{corollary}[Sample complexity]\label{cor:sample_complexity}
    Under the hypotheses of Theorem~\ref{thm:rvi_main}(ii), let $K_0$ be the smallest integer satisfying \eqref{eq:rvi_stepsize_cond}, so that $K_0=O(K^2+K^6(1+D)^2\log(2|\cS||\cA|))$. There is a universal constant $C$ such that, for every $\eps>0$, the greedy policy $\pi_{Q_n}$ satisfies $\bE\sqbr{\Delta_n}\leq\eps$ whenever
    \begin{align*}
        n\geq C\br{\frac{K^6(1+D)^2\log(2|\cS||\cA|)}{\Delta_{\rm pol}\eps}+K_0\br{1+\frac{\cE(Q_0)^2}{\Delta_{\rm pol}\eps}}},
    \end{align*}
    where $\cE(Q_0)\leq\tfrac12(\norm{Q_0}_{\rm sp}+1+3D)$. 
\end{corollary}

%% file: tex_files/km.tex
\section{Stochastic KM Iterations for Linearly Regular Maps}
\label{sec:km}

This section studies stochastic KM iterations under two hypotheses: linear regularity and a residual-dependent growth of the noise. Throughout, $(\cX,\norm{\cdot})$ is a finite-dimensional real normed vector space and $T$ is a non-expansive map on it. We define its fixed-point set, residual, and fixed-point distance below. Section~\ref{sec:rvi_rates} verifies the hypotheses for the centered Bellman operator by taking $\cX=\bR^{\cS\times\cA}$ with the sup-norm and $T=\bar H$.

\subsection{Setup and Assumptions}
\label{subsec:km_setup}

Let $(\cX, \norm{\cdot})$ be a $d$-dimensional real normed vector space. Throughout this section, we consider the operator $T: \cX \to \cX$ that satisfies the following assumption.

\begin{assumption}[Non-expansive and linearly regular]\label{assum:operator}
    (i) $T: \cX \to \cX$ be non-expansive, that is, $\norm{Tx - Ty} \leq \norm{x-y}$ for all $x, y \in \cX$. 
    
    (ii) The set of fixed points
    \begin{align*}
        \fix(T) = \flbr{x \in \cX : Tx = x}
    \end{align*}
    is non-empty and $T$ is linearly regular (Definition~\ref{def:linear_regular}), that is, there exists $\kappa \in (0,\infty)$ such that 
    \begin{align*}
        \inf_{x\ust \in \fix(T)}{\norm{x - x\ust}} \leq \kappa \norm{x - Tx}, \quad \forall x \in \cX.
    \end{align*}
\end{assumption}
For convenience, we denote $\cE(x):= \inf_{x\ust \in \fix(T)}{\norm{x - x\ust}}$. $\fix(T)$ is closed, since $T$ is continuous, but it need not be either convex or a singleton. The reverse inequality of linear regularity is always true by non-expansiveness.~Fix $x\ust \in \fix(T)$ arbitrarily.~For $x \in \cX$, non-expansiveness gives $\norm{x - Tx} \leq \norm{Tx - T x\ust} + \norm{x\ust - x} \leq 2\norm{x - x\ust}$. Taking the infimum over $x\ust$,
\begin{align}
    \norm{x - Tx} \leq 2\cE(x), \qquad x \in \cX. \label{eq:res_le_dist}
\end{align}

Linear regularity is a weaker property than contraction. To see this, suppose that $T$ is a $\gamma$-contraction, $\gamma \in (0,1)$, with unique fixed point $x^\star$. Then, $\cE(x) = \|x - x^\star\| \leq \|x - Tx\|+ \|Tx - Tx^\star\| \leq \norm{x - Tx} + \gamma\|x - x^\star\|$. Thus, every contractive operator is linearly regular with $\kappa= \frac{1}{1-\gamma}$. 

However, not all linearly regular operators are contractive. Consider the reflection operator $T(x) = -x$ on $\cX = \bR$. Since $\abs{Tx - Ty} = \abs{x - y}$, it is non-expansive and not a contraction. Its fixed-point set is $\{0\}$, and $\cE(x) = \abs{x}$ while $\norm{x - Tx} = \abs{x - (-x)} = 2\abs{x}$, so $T$ is linearly regular with $\kappa = 1$. The following example shows that linear regularity is not a consequence of non-expansiveness either.

\begin{example}\label{ex:no_error_bound}
    On $\cX = \bR$ consider the piecewise affine map
    \begin{align*}
        T(x) =
        \begin{cases}
            x, & x \leq 0, \\
            0, & 0 \leq x \leq 1, \\
            x - 1, & x \geq 1.
        \end{cases}
    \end{align*}
    Each piece has slope $0$ or $1$, and the pieces agree at the breakpoints, so $T$ is continuous and non-expansive, and $\fix(T) = (-\infty, 0]$. For $x \geq 1$, the residual is $\norm{x - Tx} = 1$ while $\cE(x) = x$. So, there exists no finite $\kappa$ such that $\cE(x) \leq \kappa \norm{x - Tx}$ for all $x$.
\end{example}

We study the stochastic KM iteration
\begin{align}
    x_n = (1-\alpha_n) x_{n-1} + \alpha_n \br{T x_{n-1} + M_n}, \qquad n \geq 1, \label{eq:km_iter}
\end{align}
where $x_0 \in \cX$ and the stepsizes $\{\alpha_n\}_{n \geq 1}$ are deterministic and $\{M_n\}_{n\geq1}$ is adapted to a filtration $\{\cF_n\}_{n \geq 0}$ with $x_n$ being $\cF_n$-measurable. The noise is allowed to grow with the current residual $\norm{x_{n-1} - Tx_{n-1}}$.

\begin{assumption}[Martingale difference noise with residual-dependent variance]\label{assum:noise}
    There is a constant $\sigma \geq 0$ such that, almost surely, for every $n \geq 1$,
    \begin{align*}
        \bE\sqbr{M_n \mid \cF_{n-1}} = 0, \qquad \bE\sqbr{\norm{M_n}^2 \mid \cF_{n-1}} \leq \sigma^2 \br{1 + \norm{x_{n-1} - T x_{n-1}}^2}.
    \end{align*}
\end{assumption}

We verify Assumption~\ref{assum:noise} for RVI $Q$-learning in Section~\ref{subsec:rvi_noise}. Let $\nu$ be any finite constant such that
\begin{align}
    \bE\norm{\sum_{i}{U_i}}^2 \leq \nu \sum_i{\bE\norm{U_i}^2} \label{eq:mds_const}
\end{align}
for every finite $\cX$-valued martingale difference sequence $\{U_i\}$. Such a constant exists by equivalence with a Euclidean norm. Lemma~\ref{lem:mds_sup} provides the admissible choice $\nu=2e\log(2d)$ for the sup-norm on $\bR^d$ when $d\geq2$ (and $\nu=1$ suffices when $d=1$). 

We consider polynomial stepsize sequences, formalized in the following assumption.
\begin{assumption}[Stepsizes]\label{assum:stepsize}
    The stepsizes are given by
    \begin{align}
        \alpha_n = c(n+K_0)^{-a}, \qquad 0<a\leq1,\quad c>0,\quad K_0\in\{1,2,\ldots\},
        \label{eq:stepsize}
    \end{align}
    for every $n\geq1$. Also,
    \begin{align}
        c K_0^{1-a} \geq 2^8\kappa^2, \qquad \alpha_1 = \frac{c}{(K_0+1)^a} \leq \frac{1}{\max\flbr{2, 2^8 \nu \sigma^2 (2^7\kappa^2+2)^2}}. \label{eq:stepsize_cond}
    \end{align}
\end{assumption}
For convenience, we write $L := 2^7\kappa^2$.

\subsection{The Residual Bound}
\label{subsec:km_main}

\begin{theorem}[Mean-squared residual bound]\label{thm:km_main}
    Let the operator $T$ satisfy Assumption~\ref{assum:operator}, and consider \eqref{eq:km_iter} under Assumptions~\ref{assum:noise} and \ref{assum:stepsize}. Then, for every $n \geq 0$,
    \begin{align}
        \bE[\norm{x_n - T x_n}^2] \leq 25\cE(x_0)^2 \exp\br{-\frac{\log(4)}{L + 1/2}\sum_{k=1}^n\alpha_k} + 45\nu\sigma^2(L+2)^2\frac{c}{(n+K_0+1)^a}. \label{eq:km_main}
    \end{align}
\end{theorem}

The first term of \eqref{eq:km_main} is the initial error, reduced by at least a factor of four once per block of iterations whose stepsizes sum to $L$, so that it is damped by a factor determined by the accumulated stepsize $\sum_{k=1}^n\alpha_k$. The second term is the noise accumulated over the single most recent such block. The noise of every earlier block has a negligible effect due to exponential decay.

\begin{remark}[Second term dominates]\label{rem:noise_dominates}
    For $a \in (0,1)$, by \eqref{eq:A_n_poly} in Lemma~\ref{lem:stepsize_sums},
    \begin{align*}
        \sum_{k=1}^n\alpha_k \geq \frac{c}{1-a}[(n+K_0+1)^{1-a} - (K_0+1)^{1-a}].
    \end{align*}
    Hence, for every $a \in (0,1)$, $\exp\br{-\frac{\log(4)}{L + 1/2}\sum_{k=1}^n\alpha_k}$ decays faster than $(n+K_0)^{-a}$. For $a = 1$, by \eqref{eq:A_n_harmonic} in Lemma~\ref{lem:stepsize_sums},
    \begin{align*}
        \sum_{k=1}^n\alpha_k \geq c\log\br{\frac{n+K_0+1}{K_0+1}}.
    \end{align*}
    If $T=I$, the residual is zero at every point. Otherwise, \eqref{eq:res_le_dist} and linear regularity imply $\kappa\geq1/2$, and hence $L\geq32$. The preceding bound and $c\geq2L$ then give $c\log(4)/(L+1/2)\geq64\log(4)/(32+1/2)>1$, so the exponential term decays faster than $1/(n+K_0)$. Thus the mean-squared residual is $O(n^{-a})$ for $0<a\leq1$.
\end{remark}

\begin{remark}[Removing the offset condition]\label{rem:offset}
    For $a\in(0,1)$, the exponent $a$ for the mean-squared residual also holds for every fixed $c>0$ and integer $K_0\geq1$ with $\alpha_1\leq1$, without imposing \eqref{eq:stepsize_cond}. For $a=1$, the same holds provided $c\geq2L$. For the proof, choose a deterministic $n_0$ for which $K_0' := K_0 + n_0$ satisfies \eqref{eq:stepsize_cond}, and apply Theorem~\ref{thm:km_main} from time $n_0$ onward. The proof uses the initial point only through the second moment of its distance to $\fix(T)$, so it applies verbatim to the random point $x_{n_0}$ with $\cE(x_0)^2$ replaced by $\bE[\cE(x_{n_0})^2]$ and $\sum_{k=1}^n\alpha_k$ by $\sum_{k=n_0+1}^n\alpha_k$. This second moment is finite. To see this, fix $x\ust \in \fix(T)$ with $\norm{x_0 - x\ust} = \cE(x_0)$. \eqref{eq:km_iter}, \eqref{eq:res_le_dist} and Minkowski's inequality give $\bE[\norm{x_k - x\ust}^2]^{\frac12} \leq (1+2\sigma\alpha_k)\bE[\norm{x_{k-1} - x\ust}^2]^{\frac12} + \sigma\alpha_k$, from which we obtain
    \begin{align*}
        \bE[\cE(x_{n_0})^2] \leq \sqbr{\br{\cE(x_0) + \tfrac12}e^{2\sigma \sum_{k=1}^{n_0}\alpha_k} - \tfrac12}^2.
    \end{align*}
    The rate $O(n^{-a})$ for the mean-squared residual therefore holds for any such fixed offset, with a possibly large transient constant.
\end{remark}
\subsection{Proof of the Residual Bound}
\label{subsec:km_proof}

\subsubsection{Contraction over a Block}
\label{subsec:block_regularity}

Define deterministic block endpoints by $t_0 := 0$ and
\begin{align}
    t_{j+1} := \min\flbr{b > t_j : \sum_{k=t_j+1}^{b}{\alpha_k} \geq L}, \qquad j \geq 0. \label{def:blocks}
\end{align}
Since $\sum_{k=1}^{\infty}\alpha_k=\infty$, every tail sum diverges, so the endpoints in \eqref{def:blocks} are finite. By \eqref{eq:stepsize_cond}, for each $j \geq 0$, $\sum_{k=t_j+1}^{t_{j+1}}\alpha_k \in [L,L+1/2]$.

\begin{lemma}\label{lem:block_regularity}
    Let $T:\cX\to\cX$ satisfy Assumption~\ref{assum:operator}. For any integer $m\geq0$ and any $y_m\in\cX$, consider the noiseless KM iteration
    \begin{align}
        y_k = (1-\alpha_k)y_{k-1} + \alpha_k T y_{k-1}, \qquad k > m, \label{eq:exact_km}
    \end{align}
    where the stepsizes satisfy the hypotheses of Theorem~\ref{thm:km_main}. Its distance to the fixed-point set is non-increasing, i.e.,
    \begin{align}
        \cE(y_k) \leq \cE(y_{k-1}) \leq \cE(y_m), \qquad k > m. \label{eq:fejer}
    \end{align}
    Moreover, if $m=t_j$, then
    \begin{align}
        \cE(y_{t_{j+1}}) \leq \frac{1}{4\sqrt{\pi}}\cE(y_{t_j}). \label{eq:block_contraction}
    \end{align}
\end{lemma}

\begin{proof}
    Fix $m\geq0$, $y_m\in\cX$, and $x\ust\in\fix(T)$, and set $R:=\norm{y_m-x\ust}$. If $R=0$, then $y_k=x\ust$ for every $k\geq m$, so the residual bound is immediate. Suppose $R>0$ and define $C:=\flbr{z\in\cX:\norm{z-x\ust}\leq R}$. This is a non-empty, bounded, closed convex subset of the normed space $\cX$, with $y_m\in C$ and $\operatorname{diam}(C)\leq2R$. For every $z\in C$, non-expansiveness and $Tx\ust=x\ust$ give
    \begin{align*}
        \norm{Tz-x\ust}=\norm{Tz-Tx\ust}\leq\norm{z-x\ust}\leq R.
    \end{align*}
    Thus $T(C)\subseteq C$, and the restriction $T|_C:C\to C$ is non-expansive. Since $\alpha_k\in[0,1]$, convexity of $C$ and induction in \eqref{eq:exact_km} show that $y_k\in C$ for every $k\geq m$.

    To apply \citet[Theorem 1]{cominetti2014rate}, set $u_\ell:=y_{m+\ell}$ for $\ell\geq0$ and $\beta_\ell:=\alpha_{m+\ell}$ for $\ell\geq1$. Then $u_0=y_m\in C$ and
    \begin{align*}
        u_\ell=(1-\beta_\ell)u_{\ell-1}+\beta_\ell (T|_C)(u_{\ell-1}), \qquad \ell\geq1.
    \end{align*}
    These are precisely the KM iterates of a non-expansive self-map of the bounded convex set $C$, with relaxation parameters $\beta_\ell\in[0,1]$, as required by that theorem. For any integer $b>m$, applying it at $N=b-m\geq1$ yields
    \begin{align*}
        \norm{y_b-Ty_b}
        &\leq\frac{\operatorname{diam}(C)}{\sqrt{\pi\sum_{\ell=1}^{N}\beta_\ell(1-\beta_\ell)}}
        \leq\frac{2\norm{y_m-x\ust}}{\sqrt{\pi\sum_{k=m+1}^{b}\alpha_k(1-\alpha_k)}}.
    \end{align*}
    The denominator is positive and independent of $x\ust$. Taking the infimum over $x\ust\in\fix(T)$, we have that
    \begin{align}
        \norm{y_b - T y_b} \leq \frac{2 \cE(y_m)}{\sqrt{\pi \sum_{k=m+1}^{b}{\alpha_k(1-\alpha_k)}}}, \qquad b > m. \label{eq:km_det}
    \end{align}

    For every $x\ust\in\fix(T)$ and $k>m$, the triangle inequality and non-expansiveness give
    \begin{align*}
        \norm{y_k-x\ust}
        &\leq(1-\alpha_k)\norm{y_{k-1}-x\ust}+\alpha_k\norm{Ty_{k-1}-Tx\ust}
        \leq\norm{y_{k-1}-x\ust}.
    \end{align*}
    Taking the infimum over $x\ust$, and iterating proves \eqref{eq:fejer}.

    For $m=t_j$ and $b=t_{j+1}$, minimality of $b$ gives $\sum_{k=m+1}^{b-1}\alpha_k<L$. Together with $\alpha_b\leq1/2$, this proves the asserted block-sum bounds. Also,
    \begin{align}
        \sum_{k=t_j+1}^{t_{j+1}}\alpha_k(1-\alpha_k)
        \geq\frac12\sum_{k=t_j+1}^{t_{j+1}}\alpha_k
        \geq\frac L2. \label{eq:regularity_time}
    \end{align}
    Finally, linear regularity, \eqref{eq:km_det}, and \eqref{eq:regularity_time} imply
    \begin{align*}
        \cE(y_b)\leq\kappa\norm{y_b-Ty_b}
        \leq\frac{2\kappa}{\sqrt{\pi L/2}}\cE(y_m)
        =\frac{1}{4\sqrt{\pi}}\cE(y_m)
        \leq\frac14\cE(y_m).
    \end{align*}
    This proves \eqref{eq:block_contraction}.\qedhere
\end{proof}

\subsubsection{Tracking a Block with Martingale Noise}
\label{subsec:block_lemma}

\begin{lemma}[Block estimate]\label{lem:block}
    Let $m = t_j$ and $b = t_{j+1}$, and put $\eta_j := \sqrt{\nu} \sigma (L+2) \sqrt{\alpha_{m+1}}$. If $\eta_j \leq 1/16$, then
    \begin{align}
        \max_{m \leq k \leq b}{\bE[\cE(x_k)^2]} &\leq \frac{48}{31}\bE[\cE(x_m)^2] + \frac{64}{31}\eta_j^2, \label{eq:block_max} \\
        \bE[\cE(x_b)^2] &\leq \frac18\bE[\cE(x_m)^2] + \frac{80}{93}\eta_j^2. \label{eq:block_end}
    \end{align}
    Bound~\eqref{eq:block_max} also holds with $b$ replaced by any $n$ with $m \leq n < t_{j+1}$.
\end{lemma}

\begin{proof}
    First, note that $\norm{Tx} \leq \norm{x} + \norm{T0}$. Fix $x\ust \in \fix(T)$. By Assumption~\ref{assum:noise} and \eqref{eq:res_le_dist}, $\bE[\norm{M_k}^2 \mid \cF_{k-1}] \leq \sigma^2(1 + 4\norm{x_{k-1}-x\ust}^2)$. So, $\bE\norm{x_{k-1}}^2 < \infty$ implies $\bE\norm{x_k}^2 < \infty$ by \eqref{eq:km_iter}. Since $x_0$ is deterministic, induction gives $\bE\norm{x_k}^2 < \infty$ for every $k$. Thus, $\bE[\cE(x_k)^2] < \infty$, and $D_{\max} := \max_{m \leq k \leq b}{\bE[\cE(x_k)^2]} < \infty$.
    
    Run the noiseless comparison iteration \eqref{eq:exact_km} from the random point $y_m := x_m$, and reset, on this block only, the averaged noise process
    \begin{align*}
        \xi_m := 0, \qquad \xi_k := (1-\alpha_k)\xi_{k-1} + \alpha_k M_k, \qquad z_k := x_k - \xi_k, \quad k > m.
    \end{align*}
    Write $\Phi_k := (1-\alpha_k)I + \alpha_k T$, where $I$ is the identity map on $\cX$. The map $\Phi_k$ is non-expansive as a convex combination of non-expansive maps, and the noise cancels in $z_k$:
    \begin{align*}
        z_k = (1-\alpha_k)z_{k-1} + \alpha_k T x_{k-1} = \Phi_k z_{k-1} + \alpha_k\br{T x_{k-1} - T z_{k-1}},
    \end{align*}
    so $\{z_k\}$ is an inexact KM iteration with per-step error at most $\alpha_k\norm{x_{k-1}-z_{k-1}} = \alpha_k\norm{\xi_{k-1}}$. Since $z_m = y_m$ and $\Phi_k$ is non-expansive, $\norm{z_k - y_k} \leq \norm{z_{k-1}-y_{k-1}} + \alpha_k\norm{\xi_{k-1}}$, and therefore, pathwise,
    \begin{align}
        &\norm{z_k - y_k} \leq \sum_{i=m+1}^{k}{\alpha_i\norm{\xi_{i-1}}} \notag\\
        \implies &\norm{x_k - y_k} \leq \norm{x_k - z_k} + \sum_{i=m+1}^{k}{\alpha_i\norm{\xi_{i-1}}} \leq \norm{\xi_k} + \sum_{i=m+1}^{k}{\alpha_i\norm{\xi_{i-1}}}, \quad m \leq k \leq b. \label{eq:pathwise_tracking}
    \end{align}
    
    Unrolling the recursion for $\xi_k$ from $\xi_m = 0$ gives $\xi_k = \sum_{i=m+1}^{k}{w_{i,k}M_i}$ with the deterministic weights $w_{i,k} := \alpha_i\prod_{\ell=i+1}^{k}{(1-\alpha_\ell)}$. Since the stepsizes are non-increasing and at most $1/2$, induction on $k$ gives
    \begin{align}
        \sum_{i=m+1}^{k}w_{i,k}^2 = (1-\alpha_k)^2\sum_{i=m+1}^{k-1}w_{i,k-1}^2+\alpha_k^2 \leq (1-\alpha_k)^2\frac{\alpha_{m+1}}{2-\alpha_{m+1}}+\alpha_k^2 \leq \frac{\alpha_{m+1}}{2-\alpha_{m+1}} \leq \frac23\alpha_{m+1}. \label{eq:weight_sq}
    \end{align}
    The induction starts with an empty sum at $k=m$; the second inequality follows from $\alpha_k/(2-\alpha_k)\leq\alpha_{m+1}/(2-\alpha_{m+1})$. The sequence $\{w_{i,k}M_i\}_{i}$ is a martingale difference sequence, so \eqref{eq:mds_const} applies.~Combining it with \eqref{eq:weight_sq}, Assumption~\ref{assum:noise}, the tower property of conditional expectation and \eqref{eq:res_le_dist},
    \begin{align*}
        \bE\norm{\xi_k}^2 \leq \nu\sum_{i=m+1}^{k}{w_{i,k}^2\,\bE\norm{M_i}^2}
        \leq \frac23\nu\alpha_{m+1}\sigma^2\br{1+4D_{\max}},
        \qquad m\leq k\leq b,
    \end{align*}
    since each index $i-1$ lies in $[m,b]$. Applying weighted Cauchy--Schwarz to \eqref{eq:pathwise_tracking} gives
    \begin{align*}
        \bE\norm{x_k-y_k}^2
        &\leq \br{1+\sum_{i=m+1}^{k}\alpha_i}
        \br{\bE\norm{\xi_k}^2+\sum_{i=m+1}^{k}\alpha_i\bE\norm{\xi_{i-1}}^2}.
    \end{align*}
    Since $1+\sum_{i=m+1}^{b}\alpha_i\leq L+3/2\leq L+2$, we obtain
    \begin{align}
        \max_{m\leq k\leq b}\bE\norm{x_k-y_k}^2
        \leq \frac23\eta_j^2\br{1+4D_{\max}}. \label{eq:tracking_L2}
    \end{align}
    The distance to a non-empty set is $1$-Lipschitz, so $\cE(x_k)\leq\cE(y_k)+\norm{x_k-y_k}$. By \eqref{eq:fejer} and $y_m=x_m$, weighted Cauchy--Schwarz and \eqref{eq:tracking_L2} give
    \begin{align*}
        \bE[\cE(x_k)^2] \leq \frac32\bE[\cE(x_m)^2]+3\bE\norm{x_k-y_k}^2 \leq  \frac32\bE[\cE(x_m)^2]+2\eta_j^2\br{1+4D_{\max}},
        \qquad m\leq k\leq b.
    \end{align*}
    The right-hand side does not depend on $k$, so $(1-8\eta_j^2)D_{\max}\leq\frac32\bE[\cE(x_m)^2]+2\eta_j^2$; this rearrangement is legitimate because $D_{\max}$ is finite. Since $\eta_j\leq1/16$ gives $1-8\eta_j^2\geq31/32$, we obtain
    \begin{align*}
        D_{\max} \leq \frac{\frac32\bE[\cE(x_m)^2]+2\eta_j^2}{1-8\eta_j^2} \leq \frac{48}{31}\bE[\cE(x_m)^2]+\frac{64}{31}\eta_j^2,
    \end{align*}
    which is \eqref{eq:block_max}. At the block endpoint, \eqref{eq:block_contraction} gives $\cE(y_b)^2\leq\cE(x_m)^2/(16\pi)$ pathwise. Using weighted Cauchy--Schwarz with coefficients $5$ and $5/4$, and substituting the first bound on $D_{\max}$ above into \eqref{eq:tracking_L2}, yields
    \begin{align*}
        \bE[\cE(x_b)^2]
        &\leq \frac{5}{16\pi}\bE[\cE(x_m)^2]+\frac54\bE\norm{x_b-y_b}^2\\
        &\leq \frac{5}{16\pi}\bE[\cE(x_m)^2]+\frac56\eta_j^2\br{1+4D_{\max}}\\
        &\leq \br{\frac{5}{16\pi}+\frac{5\eta_j^2}{1-8\eta_j^2}}\bE[\cE(x_m)^2]
            +\frac{5\eta_j^2}{6(1-8\eta_j^2)}\\
        &\leq \br{\frac{5}{48}+\frac{5}{248}}\bE[\cE(x_m)^2]+\frac{80}{93}\eta_j^2\\
        &\leq \frac18\bE[\cE(x_m)^2]+\frac{80}{93}\eta_j^2.
    \end{align*}
    We used $\pi>3$ and $5/48+5/248=185/1488<1/8$. This is \eqref{eq:block_end}, and the same argument for \eqref{eq:block_max} applies to any partial block.~This completes the proof.\qedhere
\end{proof}

\subsubsection{Completing the Proof}
\label{subsec:km_completion}

\begin{proof}[\textbf{Proof of Theorem~\ref{thm:km_main}}]
    We first verify the hypothesis of Lemma~\ref{lem:block}. By \eqref{eq:stepsize_cond}, for every $j \geq 0$,
    \begin{align}
        \eta_j^2 = \nu \sigma^2 (L+2)^2\alpha_{t_j+1} \leq \frac{\nu \sigma^2 (L+2)^2}{\max\flbr{2,2^8\nu \sigma^2 (L+2)^2}} \leq 2^{-8}, \qquad\mbox{that is,}\qquad \eta_j \leq 2^{-4}. \label{eq:eta_small}
    \end{align}
    Next, we show that the blocks at most double on the shifted clock $n \mapsto n + K_0$. Applying \eqref{eq:block_length} in Lemma~\ref{lem:stepsize_sums} with $m = t_j$, the stepsizes indexed by $t_j+1, \ldots, 2t_j+K_0$ already sum to at least $L$, so by the minimality in \eqref{def:blocks},
    \begin{align}
        t_{j+1} \leq 2t_j + K_0, \qquad\mbox{hence}\qquad t_{j+1} + K_0 \leq 2\br{t_j + K_0}. \label{eq:block_doubling}
    \end{align}
    This uses the first condition in \eqref{eq:stepsize_cond}, and it is a statement about the stepsize sequence alone, so it is valid for every $j$. By \eqref{eq:ratio} in Lemma~\ref{lem:stepsize_sums} together with \eqref{eq:block_doubling},
    \begin{align}
        \alpha_{t_j+1} \leq 2^a \alpha_{n+1} \leq 2\alpha_{n+1}, \qquad t_j \leq n \leq t_{j+1}. \label{eq:stepsize_ratio}
    \end{align}
    
    Set $e_j := \bE[\cE(x_{t_j})^2]$ and $q:=1/8$. Thus, $e_0=\cE(x_0)^2$. By \eqref{eq:eta_small}, Lemma~\ref{lem:block} applies to every block and gives $e_{j+1}\leq q e_j+\frac{80}{93}\eta_j^2$, from which we obtain
    \begin{align*}
        e_j\leq q^j\cE(x_0)^2+\frac{80}{93}\sum_{i=0}^{j-1}q^{j-1-i}\eta_i^2.
    \end{align*}
    Iterating \eqref{eq:block_doubling} gives $t_j+1+K_0\leq2^{j-i}(t_i+1+K_0)$ for $i<j$, so
    \begin{align*}
        \eta_i^2=\eta_j^2\br{\frac{t_j+1+K_0}{t_i+1+K_0}}^a
        \leq\eta_j^2\,2^{j-i},
    \end{align*}
    and, substituting $\ell=j-i$,
    \begin{align*}
        \frac{80}{93}\sum_{i=0}^{j-1}q^{j-1-i}\eta_i^2 \leq \frac{160}{93}\eta_j^2\sum_{\ell=1}^{\infty}(2q)^{\ell-1} =\frac{160}{93(1-1/4)}\eta_j^2
        =\frac{640}{279}\eta_j^2.
    \end{align*}
    Here $2q<1$. Thus, the contributions of earlier blocks form a convergent geometric sum, and $e_j\leq q^j\cE(x_0)^2+\frac{640}{279}\eta_j^2$. For $t_j\leq n<t_{j+1}$, bound~\eqref{eq:block_max} together with \eqref{eq:stepsize_ratio} gives
    \begin{align}
        \bE[\cE(x_n)^2]
        &\leq\frac{48}{31}e_j+\frac{64}{31}\eta_j^2 \leq\frac{48}{31}q^j\cE(x_0)^2+\frac{16192}{2883}\eta_j^2 \leq\frac{48}{31}4^{-j}\cE(x_0)^2 + \frac{32384}{2883}\nu\sigma^2(L+2)^2\alpha_{n+1}.
        \label{eq:rms_dn_bound}
    \end{align}
    We used $q\leq1/4$. It remains to express $j$ in terms of the accumulated stepsize. Every completed block contributes at most $L+1/2$ and the current partial block contributes less than $L$, so $\sum_{k=1}^n\alpha_k<j(L+1/2)+L\leq(j+1)(L+1/2)$, that is, $j\geq\frac{1}{L+1/2}\sum_{k=1}^n\alpha_k-1$, and therefore
    \begin{align*}
        \frac{192}{31}4^{-j}
        \leq\frac{768}{31}\exp\br{-\frac{\log(4)}{L+1/2}\sum_{k=1}^n\alpha_k}
        \leq25\exp\br{-\frac{\log(4)}{L+1/2}\sum_{k=1}^n\alpha_k}.
    \end{align*}
    Multiplying \eqref{eq:rms_dn_bound} by $4$ and using \eqref{eq:res_le_dist}, $129536/2883<45$, and $\alpha_{n+1}=c(n+K_0+1)^{-a}$ proves \eqref{eq:km_main}. For $n=0$ the claim is immediate from $\norm{x_0-Tx_0}^2\leq4\cE(x_0)^2$.\qedhere
\end{proof}

%% file: tex_files/rvi_rates.tex
\section{Proofs of the Main Results}
\label{sec:rvi_rates}

This section proves Theorem~\ref{thm:rvi_main} and Corollary~\ref{cor:sample_complexity}. We begin by introducing an auxiliary iteration governed by the centered Bellman operator, which is non-expansive in the sup-norm. Then, we show that analyzing the auxiliary iteration is sufficient to establish the convergence rate of RVI Q-learning. Subsequently, this section verifies the two hypotheses needed to apply Theorem~\ref{thm:km_main} to the auxiliary iteration, i.e., linear regularity and residual-dependent growth of the sampling noise. We then complete the proofs.

\subsection{The Centered Bellman Operator and Its Equivalence to the RVI Map}
\label{subsec:aux_map}

$H$ is non-expansive in the span seminorm, which is not a norm, and in general, $H$ is not non-expansive with respect to the sup-norm, because the normalization term $f(Q)\bo$ depends on $Q$. The obstruction disappears once that term is replaced by a constant.~To this end, define the centered Bellman operator $\bar{H}$ as
\begin{align} \label{def:aux_map}
    \bar{H} Q := BQ - J\ust \bo,\quad \forall Q \in \bR^{\cS \times \cA},
\end{align}
where $B$ is the Bellman operator~\eqref{def:bellman}. $\bar H$ is non-expansive with respect to the sup-norm, and the set of fixed points of $\bar H$, denoted by $\fix(\bar H)$, is exactly the set of solutions of the AROE~\eqref{eq:aroe}. Define the following auxiliary iteration:
\begin{align}
    \bar{Q}_{n+1} = \bar{Q}_n + \alpha_{n+1} \br{\bar{H} \bar{Q}_n - \bar{Q}_n + M_{n+1}(\bar{Q}_n)}, \quad n\geq 0,\label{rvi_km_iter}
\end{align}
where $\bar{Q}_0 = Q_0$. The auxiliary iterates $\{\bar{Q}_n\}$ are constructed on the same probability space and use the same samples $Y^{n+1}_{s,a}$ as \eqref{eq:rvi_update}. The value $J\ust$ is not known to the agent. Hence, this iteration cannot be used to learn $Q\ust$ and will be used only for analysis. It turns out that $\norm{Q_n - H Q_n}_{\rm sp} \leq 2 \norm{\bar{Q}_n - \bar{H} \bar{Q}_n}_{\infty}$, which is the next lemma.

\begin{lemma}\label{lem:equiv_iter}
    For every $n\geq0$, $\norm{Q_n - H Q_n}_{\rm sp} \leq 2 \norm{\bar{Q}_n - \bar{H} \bar{Q}_n}_{\infty}$.
\end{lemma}
\begin{proof}
    We first show that $\norm{Q_n - \bar{Q}_n}_{\rm sp} = 0$ using induction. The base case holds due to equality. Assume the induction hypothesis, $\bar{Q}_n = Q_n + c_n \bo$, for some $c_n \in \bR$. Then,
    \begin{align*}
        \bar{Q}_{n+1}(s,a) &= \bar{Q}_n(s,a) + \alpha_{n+1} \br{r(s,a,Y_{s,a}^{n+1}) + V_{\bar{Q}_n}(Y_{s,a}^{n+1}) - \bar{Q}_n(s,a) - J\ust} \\
        &= Q_n(s,a) + c_n + \alpha_{n+1} \br{r(s,a,Y_{s,a}^{n+1}) + V_{Q_n}(Y_{s,a}^{n+1}) - Q_n(s,a) - f(Q_n) + (f(Q_n) - J\ust)} \\
        &= Q_{n+1}(s,a) + c_n + \alpha_{n+1} (f(Q_n) - J\ust).
    \end{align*}
    Since $c_{n+1} := c_n + \alpha_{n+1} (f(Q_n) - J\ust)$ does not depend on $(s,a)$, the claim follows.

     Note that $\bar Q_n-\bar H\bar Q_n = Q_n + c_n \bo - B (Q_n + c_n \bo) + J\ust\bo = Q_n - B Q_n + J\ust \bo$, which differs from $Q_n-BQ_n$ by a multiple of $\bo$. Hence $\norm{\bar{Q}_n - \bar{H} \bar{Q}_n}_{\rm sp} = \norm{Q_n - B Q_n}_{\rm sp} = \norm{Q_n - H Q_n}_{\rm sp}$. The proof is then complete due to the fact that $\norm{\cdot}_{\rm sp} \leq 2\norm{\cdot}_\infty$.\qedhere
\end{proof}
Hence, it suffices to analyze \eqref{rvi_km_iter} in order to study the convergence behavior of $\norm{Q_n - H Q_n}_{\rm sp}$, and hence of the policy suboptimality $\Delta_n$.

\subsection{Linear Regularity}
\label{subsec:rvi_eb}

Our goal is to show that $\bar H$ is linearly regular in the sup-norm (Definition~\ref{def:linear_regular}), namely, to find a finite constant $\kappa\geq1$ such that
\begin{align*}
    \cE(Q)\leq\kappa\norm{Q-\bar H Q}_\infty,\qquad Q\in\bR^{\cS\times\cA},
\end{align*}
where $\cE(Q)=\inf_{q\in\fix(\bar H)}\norm{Q-q}_\infty$ is the distance to $\fix(\bar H)$. We first bound the span of a $Q$-function by its residual. By Lemma~\ref{lem:coercivity} in Appendix~\ref{sec:app_supporting}, every $Q\in\bR^{\cS\times\cA}$ satisfies
\begin{align*}
    \norm{Q}_{\rm sp} \leq (1+3D)\br{1+\norm{Q-\bar H Q}_{\rm sp}},
\end{align*}
and in particular $\norm{Q}_{\rm sp}\leq 1+3D$ for every $Q \in \fix(\bar H)$. Shift invariance of $\fix(\bar H)$ allows $\cE(Q)$ to be expressed using the span of $Q$. For every $u\in \bR^{\cS \times \cA}$, centering its range gives $\inf_{c\in\bR}\norm{u-c\bo}_\infty=\frac12\norm{u}_{\rm sp}$. Hence,
\begin{align}
    \cE(Q)=\frac12\inf_{z\in\fix(\bar H)}\norm{Q-z}_{\rm sp}
    \leq\frac12\br{\norm{Q}_{\rm sp} + (1+3D)}
    &\leq (1+3D)\br{1+\tfrac12\norm{Q-\bar H Q}_{\rm sp}} \notag\\
    &\leq (1+3D)\br{1+\norm{Q-\bar H Q}_\infty}. \label{eq:distance_growth}
\end{align}
The next proposition removes the additive constant from the above upperbound, with the explicit regularity constant $K$ of \eqref{def:K}.

\begin{proposition}[Linear regularity]
\label{prop:rvi_error_bound}
    Let Assumption~\ref{assum:wc} hold. Then $\bar H$ is linearly regular in the sup-norm with constant $K$, that is,
    \begin{align*}
        \cE(Q)\leq K\norm{Q-\bar H Q}_\infty,\qquad Q\in\bR^{\cS\times\cA}.
    \end{align*}
\end{proposition}

\begin{proof}
    Fix $Q\in\bR^{\cS\times\cA}$. We consider two cases, corresponding to the two terms in \eqref{def:K}.

    \emph{Case 1: $\norm{Q-\bar HQ}_\infty\geq\Delta_{\rm pol}$.}
    Then $1\leq\norm{Q-\bar HQ}_\infty/\Delta_{\rm pol}$, so \eqref{eq:distance_growth} gives
    \begin{align*}
        \cE(Q) \leq(1+3D)\br{1+\norm{Q-\bar HQ}_\infty} \leq(1+3D)\br{1+\frac{1}{\Delta_{\rm pol}}}\norm{Q-\bar HQ}_\infty \leq K\norm{Q-\bar HQ}_\infty.
    \end{align*}

    \emph{Case 2: $\norm{Q-\bar HQ}_\infty<\Delta_{\rm pol}$.}
    For $x\in\bR^\cS$ and $\pi\in\Pi_{\mathrm{SD}}$, define
    \begin{align*}
        (\bar Gx)(s,a)&:=\ol r(s,a)-J\ust+\sum_{s'\in\cS}p(s,a,s')x(s'),\\
        Fx&:=V_{\bar Gx},\qquad
        F_\pi x:=\ol r_\pi-J\ust\bo+P^\pi x,
    \end{align*}
    where $\ol r_\pi(s):=\ol r(s,\pi(s))$ and $P^\pi(s,s'):=p(s,\pi(s),s')$. Here, $\bo$ denotes the all-ones vector in $\bR^\cS$, and vector inequalities are coordinatewise. The maps $\bar G$, $F$, and $F_\pi$ are monotone and commute with constant shifts.

    Since $\bar GV_Q=\bar HQ$, $Q(s,a)\leq V_Q(s)$, $Q(s,\pi_Q(s))=V_Q(s)$, and every entry of $\bar HQ-Q$ lies in $[-\norm{Q-\bar HQ}_\infty, \norm{Q-\bar HQ}_\infty]$, we have
    \begin{align}
        FV_Q\leq V_Q+\norm{Q-\bar HQ}_\infty\bo,\qquad
        F_{\pi_Q} V_Q\geq V_Q-\norm{Q-\bar HQ}_\infty\bo.
        \label{eq:hit_greedy_drift}
    \end{align}
    Taking expectations under $\pi_Q$ and summing the second inequality in \eqref{eq:hit_greedy_drift} over $k$ steps gives
    \begin{align*}
        \frac1k\bE_s^{\pi_Q}\sqbr{\sum_{t=0}^{k-1}\ol r_{\pi_Q}(S_t)}
        \geq J\ust-\norm{Q-\bar HQ}_\infty
        +\frac{V_Q(s)-\bE_s^{\pi_Q}[V_Q(S_k)]}{k}.
    \end{align*}
    Since $V_Q$ is bounded, letting $k\to\infty$ yields $0\leq J\ust-J(\pi_Q,s)\leq\norm{Q-\bar HQ}_\infty<\Delta_{\rm pol}$ for every $s$. Hence \eqref{def:policy_gap} shows that $\pi_Q$ is optimal from every state.

    Since $F_{\pi_Q}V_Q\leq FV_Q$, \eqref{eq:hit_greedy_drift} implies $\norm{FV_Q-V_Q}_\infty\leq\norm{Q-\bar HQ}_\infty<\Delta_{\rm pol}$. Thus Lemma~\ref{lem:envelope} in Appendix~\ref{sec:app_supporting} applies with $v=V_Q$ and $\rho=\norm{Q-\bar HQ}_\infty$, and gives $w\in\bR^\cS$ such that
    \begin{align}
        Fw\leq w,\qquad V_Q\leq w\leq V_Q+\mathcal T\norm{Q-\bar HQ}_\infty\bo.
        \label{eq:hit_initial_envelope}
    \end{align}

    Let $C$ be the union of all recurrent classes of $P^{\pi_Q}$. Fix one recurrent class, and let $\mu$ be its stationary distribution, extended by zero outside that class, so that $\mu^\top P^{\pi_Q}=\mu^\top$. For $s$ in that class, the averages $\frac1k\sum_{t=0}^{k-1}(P^{\pi_Q})^t(s,\cdot)$ converge to $\mu$ as $k\to\infty$, so $J(\pi_Q,s)=\mu^\top\ol r_{\pi_Q}$, which equals $J\ust$ because $\pi_Q$ is optimal. Using $F_{\pi_Q} w\leq Fw\leq w$ and $\mu^\top P^{\pi_Q}=\mu^\top$, we get
    \begin{align*}
        w-F_{\pi_Q} w\geq0,\qquad
        \mu^\top(w-F_{\pi_Q} w)=J\ust-\mu^\top\ol r_{\pi_Q}=0.
    \end{align*}
    The strict positivity of $\mu$ on its class implies $F_{\pi_Q} w=w$ on that class. As the class was arbitrary, $F_{\pi_Q} w=w$ on $C$, and then $F_{\pi_Q} w\leq Fw\leq w$ gives $Fw=w$ on $C$.

    Define $w_0:=w$ and $w_{n+1}:=Fw_n$. By monotonicity and $Fw\leq w$, this sequence is non-increasing. It is bounded below by a sufficiently low constant shift of a fixed point of $F$: such a fixed point exists because $V_q=FV_q$ for any $q\in\fix(\bar H)$. Thus $w_n$ converges to a finite vector $h$, and continuity gives $Fh=h$ and $h\leq w$. Moreover, $w_n=w$ on $C$ for every $n$. Indeed, if this holds at step $n$, closure of $C$ under $\pi_Q$ and $w_n\leq w$ give, on $C$,
    \begin{align*}
        w=F_{\pi_Q} w_n\leq Fw_n\leq Fw=w.
    \end{align*}
    Consequently, $h=w\geq V_Q$ on $C$.

    Set $d:=V_Q-h$. By \eqref{eq:hit_greedy_drift} and $h=Fh\geq F_{\pi_Q} h$,
    \begin{align*}
        d\leq P^{\pi_Q} d+\norm{Q-\bar HQ}_\infty\bo,\qquad d\leq0\text{ on }C.
    \end{align*}
    The chain under $\pi_Q$ reaches $C$ almost surely from every state. Since $\pi_Q$ is optimal from every state, \eqref{def:finite_hitting_time} gives $\bE_s^{\pi_Q}[\tau_C]\leq\mathcal T$. Iterating the preceding inequality up to $\tau_C\wedge n$ gives
    \begin{align*}
        d(s)\leq\bE_s^{\pi_Q}[d(S_{\tau_C\wedge n})]
        +\norm{Q-\bar HQ}_\infty\bE_s^{\pi_Q}[\tau_C\wedge n].
    \end{align*}
    Since $d$ is bounded and $\tau_C$ has finite mean, letting $n\to\infty$ yields
    \begin{align*}
        d(s)\leq\bE_s^{\pi_Q}[d(S_{\tau_C})]
        +\norm{Q-\bar HQ}_\infty\bE_s^{\pi_Q}[\tau_C]
        \leq\mathcal T\norm{Q-\bar HQ}_\infty.
    \end{align*}
    Together with \eqref{eq:hit_initial_envelope} and $h\leq w$, this proves
    \begin{align*}
        V_Q-\mathcal T\norm{Q-\bar HQ}_\infty\bo\leq h\leq V_Q+\mathcal T\norm{Q-\bar HQ}_\infty\bo,
    \end{align*}
    that is, $\norm{V_Q-h}_\infty\leq\mathcal T\norm{Q-\bar HQ}_\infty$.

    Finally, set $q:=\bar Gh$. Then $V_q=Fh=h$, so $\bar Hq=\bar GV_q=\bar Gh=q$. Since
    $(\bar GV_Q-\bar Gh)(s,a)=\sum_{s'}p(s,a,s')(V_Q(s')-h(s'))$,
    \begin{align*}
        \cE(Q)\leq\norm{Q-q}_\infty
        &\leq\norm{Q-\bar HQ}_\infty+\norm{\bar GV_Q-\bar Gh}_\infty\\
        &\leq\norm{Q-\bar HQ}_\infty+\norm{V_Q-h}_\infty\\
        &\leq(1+\mathcal T)\norm{Q-\bar HQ}_\infty
        \leq K\norm{Q-\bar HQ}_\infty.
    \end{align*}
    This completes the proof in both cases.\qedhere
\end{proof}

\subsection{Sampling Noise}
\label{subsec:rvi_noise}

We next verify Assumption~\ref{assum:noise}.

\begin{proposition}\label{prop:rvi_noise}
    Let Assumption~\ref{assum:wc} hold.~For every $n\geq1$ and every $\cF_{n-1}$-measurable $Q\in\bR^{\cS\times\cA}$ with $\bE\norm{Q}_\infty^2<\infty$, almost surely,
    \begin{align*}
        \bE[M_n(Q)\mid\cF_{n-1}]=0,\qquad
        \bE[\norm{M_n(Q)}_\infty^2\mid\cF_{n-1}]\leq8(1+3D)^2\br{1+\|Q - \bar H Q\|_\infty^2}.
    \end{align*}
\end{proposition}

\begin{proof}
    Conditional sampling gives $\bE[M_n(Q)\mid\cF_{n-1}]=0$ for every $\cF_{n-1}$-measurable $Q$. For each coordinate,
    \begin{align*}
        \abs{M_n(Q)(s,a)}
        &\leq\abs{r(s,a,Y^n_{s,a})-\sum_{s'\in \cS}{p(s,a,s') r(s,a,s')}} +\abs{V_Q(Y^n_{s,a})-\sum_{s'}p(s,a,s')V_Q(s')} \\
        &\leq 1 + \norm{V_Q}_{\rm sp} \\
        &\stackrel{(a)}{\leq} 1 + 3D\br{1+\norm{Q-\bar H Q}_{\rm sp}} \\
        &\stackrel{(b)}{\leq} 2 (1+3D)\br{1+\|Q - \bar H Q\|_\infty} \\
        &\stackrel{(c)}{\leq} 2\sqrt2\,(1+3D) \sqrt{1+\|Q - \bar H Q\|_\infty^2},
    \end{align*}
    where $(a)$ follows from Lemma~\ref{lem:coercivity}, $(b)$ follows from the fact that $\|\cdot\|_{\rm sp} \leq 2\|\cdot\|_\infty$, and $(c)$ follows from $(1+t)^2\leq2(1+t^2)$. This pathwise bound implies the conditional second-moment bound. \qedhere
\end{proof}

\subsection{Proofs of the Rates and Sample Complexity}
\label{subsec:rvi_rate_bounds}

We now apply Theorem~\ref{thm:km_main} using Propositions~\ref{prop:rvi_error_bound} and~\ref{prop:rvi_noise}. By the discussion after \eqref{eq:greedy_loss} and Lemma~\ref{lem:equiv_iter},
\begin{align*}
    \Delta_n\leq\frac{\Delta_n^2}{\Delta_{\rm pol}}
    \leq\frac{\norm{Q_n-BQ_n}_{\rm sp}^2}{\Delta_{\rm pol}}
    \leq\frac{4}{\Delta_{\rm pol}}\norm{\bar Q_n-\bar H\bar Q_n}_\infty^2.
\end{align*}
Thus, a mean-squared sup-norm residual bound for the auxiliary iteration gives an expected policy guarantee from every initial state.

\begin{proof}[\textbf{Proof of Theorem~\ref{thm:rvi_main}}]
    Apply Theorem~\ref{thm:km_main} on $(\bR^{\cS\times\cA},\norm{\cdot}_\infty)$ with $T=\bar H$ and $x_n=\bar Q_n$. The map is non-expansive and has a non-empty fixed-point set by Section~\ref{subsec:aux_map}, and \eqref{rvi_km_iter} is the required KM iteration.~By Proposition~\ref{prop:rvi_error_bound}, the map is linearly regular with constant $K\geq1$. Proposition~\ref{prop:rvi_noise} verifies Assumption~\ref{assum:noise} with noise constant $2\sqrt2\,(1+3D)$, and the sup-norm martingale constant stated in Section~\ref{subsec:km_setup} gives \eqref{eq:mds_const} with constant $2e\log(2|\cS||\cA|)$. Substituting these three constants into \eqref{eq:stepsize_cond} yields \eqref{eq:rvi_stepsize_cond} and also gives
    \begin{align*}
        \nu\sigma^2(L+2)^2=2^6 e(2^6K^2+1)^2(1+3D)^2\log(2|\cS||\cA|).
    \end{align*}
    Since $\bar Q_0=Q_0$, the initial distance in Theorem~\ref{thm:km_main} is $\cE(Q_0)$. Multiplying \eqref{eq:km_main} by $4$ and using Lemma~\ref{lem:equiv_iter} gives
    \begin{align*}
        \bE\norm{Q_n-BQ_n}_{\rm sp}^2 &\leq 100\cE(Q_0)^2\exp\br{-\frac{\log(4)}{2^7K^2+\frac12}\sum_{k=1}^n\alpha_k} \\
        &\quad + 45\cdot2^8 e(2^6K^2+1)^2(1+3D)^2\log(2|\cS||\cA|)\,\alpha_{n+1}.
    \end{align*}
    Dividing by $\Delta_{\rm pol}$ and using the preceding policy-loss bound and $\alpha_{n+1}=c(n+K_0+1)^{-a}$ proves \eqref{eq:rvi_main}.

    For part (i), let $a\in(0,1)$. In \eqref{eq:rvi_stepsize_cond}, $cK_0^{1-a}$ increases to infinity with $K_0$ and $\alpha_1=c/(K_0+1)^a$ decreases to zero, so both inequalities hold for every sufficiently large $K_0$. Since $t\mapsto t^{-a}$ is decreasing,
    \begin{align*}
        \sum_{k=1}^n\alpha_k\geq\int_{K_0+1}^{n+K_0+1}ct^{-a}\,dt
        =\frac{c}{1-a}\sqbr{(n+K_0+1)^{1-a}-(K_0+1)^{1-a}}.
    \end{align*}
    Thus, the transient in \eqref{eq:rvi_main} decays faster than every negative power of $n$, while its last term is a constant multiple of $(n+K_0+1)^{-a}\leq n^{-a}$. Hence $n^a$ times the right-hand side of \eqref{eq:rvi_main} is bounded over $n\geq1$, and its supremum is an admissible $C$.

    For part (ii), let $a=1$ and $c=2^8K^2$. The same integral comparison gives $\sum_{k=1}^n\alpha_k\geq c\log((n+K_0+1)/(K_0+1))$, so the first term of \eqref{eq:rvi_main} is at most $100\cE(Q_0)^2((K_0+1)/(n+K_0+1))^p/\Delta_{\rm pol}$. Moreover, $\alpha_{n+1}=2^8K^2/(n+K_0+1)$. Substituting both into \eqref{eq:rvi_main} gives \eqref{eq:rvi_hit_harmonic}. Since $p$ increases with $K\geq1$, we have $p\geq2^8\log(4)/(2^7+1/2)>1$.\qedhere
\end{proof}

\begin{proof}[\textbf{Proof of Corollary~\ref{cor:sample_complexity}}]
    By \eqref{eq:rvi_hit_harmonic}, $\bE[\Delta_n]\leq\eps$ whenever both of its terms are at most $\eps/2$, that is, whenever
    \begin{align*}
        n+K_0+1\geq\max\flbr{
            \frac{45\cdot2^{17} e K^2(2^6K^2+1)^2(1+3D)^2\log(2|\cS||\cA|)}{\Delta_{\rm pol}\eps},\
            (K_0+1)\br{\frac{200\cE(Q_0)^2}{\Delta_{\rm pol}\eps}}^{1/p}}.
    \end{align*}
    The first term of the maximum is $O(K^6(1+D)^2\log(2|\cS||\cA|)/(\Delta_{\rm pol}\eps))$. Since $p>1$, $(200\cE(Q_0)^2/(\Delta_{\rm pol}\eps))^{1/p}\leq1+200\cE(Q_0)^2/(\Delta_{\rm pol}\eps)$, so the second term is $O(K_0(1+\cE(Q_0)^2/(\Delta_{\rm pol}\eps)))$. This proves the sufficient condition on $n$.

    With $a=1$ and $c=2^8K^2$, the first inequality in \eqref{eq:rvi_stepsize_cond} holds for every $K_0$, and the second, in which $\alpha_1=c/(K_0+1)$, holds if
    \begin{align*}
        K_0+1\geq2^8K^2\max\flbr{2,2^{14}e(1+3D)^2(2^6K^2+1)^2\log(2|\cS||\cA|)}.
    \end{align*}
    Hence the smallest admissible $K_0$ is $O(K^2+K^6(1+D)^2\log(2|\cS||\cA|))$. The initial-distance estimate follows from \eqref{eq:distance_growth}, and each synchronous iteration uses $|\cS||\cA|$ transition samples.\qedhere
\end{proof}

%% file: tex_files/conclusion.tex
\section{Conclusion}
\label{sec:conclusion}

For each fixed finite weakly communicating MDP, we show that the expected suboptimality of the greedy policy produced by synchronous RVI $Q$-learning decreases at a rate of $O(n^{-1})$. The analysis exploits the linear regularity property of the relevant operator to obtain the convergence rate of $O(n^{-1})$ for the mean-squared residual, and the positive policy gap of finite MDPs enables the same rate for the suboptimality of the greedy policy with respect to the current $Q$ iterate.

Sharpening the current parameter dependence or proving that the dependence is unimprovable remains unaddressed. Extending the analysis to high-probability guarantees would further broaden the scope of the results. The convergence rate of asynchronous RVI $Q$-learning remains an open direction for future work.

\subsection*{Disclosure of AI Use}
The authors independently formulated the noise condition under which the conditional second moment grows with the squared residual and established that RVI Q-learning satisfies this condition (Proposition~\ref{prop:rvi_noise}). Subsequently, GPT-6 Astra (OpenAI) suggested the linear regularity condition and provided preliminary proofs supporting the improved convergence rate. AI tools were also used to assist with writing and editing the manuscript. The authors have independently verified all results, proofs, and references and take full responsibility for the content of the paper.

%% file: appendix/lin_regularity.tex
\section{Supporting Lemmas}
\label{sec:app_supporting}

This appendix proves the two lemmas used in Section~\ref{subsec:rvi_eb}. Lemma~\ref{lem:coercivity} bounds the span of a $Q$-function by its residual, and Lemma~\ref{lem:envelope} constructs the vector $w$ used in the proof of Proposition~\ref{prop:rvi_error_bound}. We use the maps $\bar G$, $F$ and $F_\pi$ and the transition matrices $P^\pi$ defined in that proof.

\begin{lemma}\label{lem:coercivity}
    Let Assumption~\ref{assum:wc} hold. For every $Q\in \bR^{\cS \times \cA}$,
    \begin{align*}
        \norm{V_Q}_{\rm sp} \leq 3D\br{1+\norm{Q-\bar H Q}_{\rm sp}},\quad
        \norm{Q}_{\rm sp} \leq (1+3D)\br{1+\norm{Q-\bar H Q}_{\rm sp}}.
    \end{align*}
    In particular, $\norm{Q}_{\rm sp}\leq 1+3D$ for every $Q \in \fix(\bar H)$.
\end{lemma}

\begin{proof}
    Decompose the residual into a scalar and a small part,
    \begin{align}
        Q - BQ = \lambda\bo + \zeta, \qquad \norm{\zeta}_\infty \leq \frac{\norm{Q-\bar H Q}_{\rm sp}}{2}, \label{eq:residual_split}
    \end{align}
    where $\lambda$ is the midpoint of the range of $Q - BQ$. The greedy policy $\pi_Q$ of Section~\ref{subsec:rvi_algo} satisfies $Q(s,\pi_Q(s)) = V_Q(s)$. Evaluating \eqref{eq:residual_split} at $(s,\pi_Q(s))$ and using \eqref{def:bellman} gives the Poisson equation with zero gain
    \begin{align}
        V_Q = P^{\pi_Q}V_Q + h, \qquad h(s) := \ol{r}\br{s,\pi_Q(s)} + \lambda + \zeta\br{s,\pi_Q(s)}, \label{eq:poisson}
    \end{align}
    where $P^{\pi}$ denotes the transition matrix induced by $\pi$. Let $\mu_{\pi_Q}$ be a stationary distribution corresponds to $P^{\pi_Q}$. Multiplying \eqref{eq:poisson} by $\mu_{\pi_Q}^\top$ gives $\mu_{\pi_Q}^\top h = 0$, that is, $\lambda = -\mu_{\pi_Q}^\top\ol{r}(\cdot,\pi_Q(\cdot)) - \mu_{\pi_Q}^\top\zeta(\cdot,\pi_Q(\cdot))$. Since $\ol{r} \in [0,1]$ and $\norm{\zeta}_\infty \leq \norm{Q-\bar H Q}_{\rm sp}/2$,
    \begin{align}
        -1-\frac{\norm{Q-\bar H Q}_{\rm sp}}{2} \leq \lambda \leq \frac{\norm{Q-\bar H Q}_{\rm sp}}{2}, \quad\mbox{and consequently}\quad \norm{h}_\infty \leq 1 + \norm{Q-\bar H Q}_{\rm sp}. \label{eq:lambda_bound}
    \end{align}
    Furthermore, $V_Q(s) \geq Q(s,\pi(s))$ for every $\pi \in \Pi_{\mathrm{SD}}$, so \eqref{eq:residual_split} also yields the family of inequalities
    \begin{align}
        V_Q \geq P^{\pi}V_Q + g_\pi, \qquad g_\pi(s) := \ol{r}\br{s,\pi(s)} + \lambda - \frac{\norm{Q-\bar H Q}_{\rm sp}}{2}, \qquad \pi \in \Pi_{\mathrm{SD}}, \label{eq:supersolution}
    \end{align}
    and $-g_\pi \leq 1 + \norm{Q-\bar H Q}_{\rm sp}$ by \eqref{eq:lambda_bound}. Unlike \eqref{eq:poisson}, which holds only for the greedy policy, \eqref{eq:supersolution} holds for every policy, so the policy used to travel between two states of $\cS_{\rm c}$ may be chosen freely.

    Let $s_-$ and $s_+$ attain the minimum and the maximum of $V_Q$ over $\cS_{\rm c}$. If $s_- = s_+$ the range over $\cS_{\rm c}$ is zero. Otherwise pick $\pi \in \Pi_{\mathrm{SD}}$ attaining $\min_{\pi'}{\bE^{\pi'}_{s_-}[\tau_{s_+}]}$, which is at most $D$ by \eqref{def:D}, and apply optional stopping to \eqref{eq:supersolution} at $\tau_{s_+}$:
    \begin{align*}
        V_Q(s_-) \geq \bE_{s_-}^{\pi}\sqbr{V_Q(S_{\tau_{s_+}})} + \bE_{s_-}^{\pi}\sqbr{\sum_{t=0}^{\tau_{s_+}-1}{g_\pi(S_t)}} \geq V_Q(s_+) - \bE_{s_-}^\pi[\tau_{s_+}]\br{1+\norm{Q-\bar H Q}_{\rm sp}},
    \end{align*}
    so that $\max_{\cS_{\rm c}}V_Q - \min_{\cS_{\rm c}}V_Q \leq D(1+\norm{Q-\bar H Q}_{\rm sp})$. For $s \in \cS_{\rm t}$, apply optional stopping to the equality \eqref{eq:poisson} at $\tau_{\cS_{\rm c}}$ under $\pi_Q$, whose expectation is at most $D$ by \eqref{def:D}:
    \begin{align*}
        \abs{V_Q(s) - \bE_s^{\pi_Q}\sqbr{V_Q\br{S_{\tau_{\cS_{\rm c}}}}}} = \abs{\bE_s^{\pi_Q}\sqbr{\sum_{t=0}^{\tau_{\cS_{\rm c}}-1}{h(S_t)}}} \leq D\norm{h}_\infty \leq D(1+\norm{Q-\bar H Q}_{\rm sp}).
    \end{align*}
    Since $S_{\tau_{\cS_{\rm c}}} \in \cS_{\rm c}$, the expectation on the left lies between the minimum and the maximum of $V_Q$ over $\cS_{\rm c}$, whence $\norm{V_Q}_{\rm sp} \leq D(1+\norm{Q-\bar H Q}_{\rm sp}) + 2D(1+\norm{Q-\bar H Q}_{\rm sp}) = 3D(1+\norm{Q-\bar H Q}_{\rm sp})$. Both stopping arguments are justified by first stopping at the hitting time truncated at $m$ and letting $m \to \infty$: $\cS$ is finite, so $V_Q$, $\ol{r}$, and $h$ are bounded, and the hitting times used have finite means under Assumption~\ref{assum:wc}.

    Finally, by \eqref{def:bellman} the range of $BQ$ is at most the range of $\ol{r}$, which is at most $1$, plus the range of $V_Q$, so
    \begin{align*}
        \norm{Q}_{\rm sp}
        &\leq \norm{BQ}_{\rm sp} + \norm{Q-\bar H Q}_{\rm sp}\\
        &\leq 1 + \norm{V_Q}_{\rm sp} + \norm{Q-\bar H Q}_{\rm sp}\\
        &\leq 1 + 3D(1+\norm{Q-\bar H Q}_{\rm sp}) + \norm{Q-\bar H Q}_{\rm sp}\\
        &= (1+3D)(1+\norm{Q-\bar H Q}_{\rm sp}).
    \end{align*}
    For $z \in \fix(\bar H)$ we have $\norm{z-\bar H z}_{\rm sp} = \norm{z-\bar H z}_\infty = 0$ by the fixed-point identity $z=\bar H z$, so $\norm{z}_{\rm sp} \leq 1+3D$.\qedhere
\end{proof}

For the next lemma and its proof, we call a vector $w\in\bR^\cS$ a \emph{super-solution} if $Fw\leq w$, equivalently, $V_{\bar Gw}\leq w$, with inequalities understood coordinatewise.

\begin{lemma}[Minimal super-solution]\label{lem:envelope}
    Let Assumption~\ref{assum:wc} hold, and let $v\in\bR^\cS$ satisfy $\norm{V_{\bar Gv}-v}_\infty\leq\rho$ for some $0\leq\rho<\Delta_{\rm pol}$. Then there is a smallest super-solution $w\geq v$, and
    \begin{align}
        0\leq w-v\leq\mathcal T\rho\bo.\label{eq:hit_envelope}
    \end{align}
\end{lemma}

\begin{proof}
    \emph{Step 1:} If $x\leq x'$, then $\bar Gx\leq\bar Gx'$, hence $V_{\bar Gx}\leq V_{\bar Gx'}$. Set
    \begin{align*}
        w_0:=v,\qquad w_{k+1}:=\max\flbr{v,V_{\bar Gw_k}},\quad k\geq0.
    \end{align*}
    (a) $\{w_k\}$ is non-decreasing: $w_1=\max\{v,V_{\bar Gv}\}\geq v=w_0$, and if $w_k\geq w_{k-1}$, then $w_{k+1}=\max\{v,V_{\bar Gw_k}\}\geq\max\{v,V_{\bar Gw_{k-1}}\}=w_k$.

    (b) $\{w_k\}$ is bounded above: fix $Q\ust\in\fix(\bar H)$, which is non-empty by Assumption~\ref{assum:wc}, and set $u:=V_{Q\ust}$, $c:=\max_{s\in\cS}\{v(s)-u(s)\}$, and $z:=u+c\bo$. Then $z\geq v$, and since $\bar Gu=\bar HQ\ust=Q\ust$,
    \begin{align*}
        V_{\bar Gz}=V_{\bar Gu}+c\bo=V_{Q\ust}+c\bo=z.
    \end{align*}
    Thus $w_0\leq z$, and $w_k\leq z$ implies $w_{k+1}\leq\max\{v,V_{\bar Gz}\}=z$.

    By (a) and (b), each $w_k(s)$ is non-decreasing in $k$ and bounded above by $z(s)$, so $w_k\to w$ for some $w\in\bR^\cS$. Since $x\mapsto\max\{v,V_{\bar Gx}\}$ is continuous,
    \begin{align}
        w=\max\flbr{v,V_{\bar Gw}}.\label{eq:hit_envelope_fixed}
    \end{align}
    Hence $w\geq v$ and $w\geq V_{\bar Gw}$. If $w'\geq v$ is a super-solution, then $w_0\leq w'$, and $w_k\leq w'$ implies $w_{k+1}\leq\max\{v,V_{\bar Gw'}\}\leq w'$, so $w\leq w'$. Thus $w$ is the smallest super-solution above $v$.

    \emph{Step 2:} Let $C_0:=\{s:w(s)=v(s)\}$, and call $a$ active at $s$ if $w(s)=(\bar Gw)(s,a)$. If $s\notin C_0$, then $w(s)>v(s)$, so \eqref{eq:hit_envelope_fixed} gives $w(s)=V_{\bar Gw}(s)=\max_a(\bar Gw)(s,a)$, and $s$ has an active action. Let $\mathcal G$ be the directed graph on $\cS$ with an edge $s\to s'$ if $p(s,a,s')>0$ for some $a$ active at $s$. We claim that every state has a path to $C_0$ in $\mathcal G$. Suppose not, and let $U\neq\varnothing$ be the set of states with no such path. Then $U\cap C_0=\varnothing$, and
    \begin{align}
        \sum_{s'\in U}p(s,a,s')=1,\qquad s\in U,\ a\text{ active at }s,\label{eq:hit_closed_U}
    \end{align}
    since an edge from $s$ to a state outside $U$ would give $s$ a path to $C_0$. Choose $\delta>0$ such that
    \begin{align*}
        \delta \leq w(s)-v(s),\quad s\in U,  \quad \text{and}\quad
        \delta \leq w(s)-(\bar Gw)(s,a), \quad s\in U,\ a \text{ not active at } s.
    \end{align*}
    The first inequality is possible because $U\cap C_0=\varnothing$, and the second inequality is possible because $w\geq V_{\bar Gw}$ and $a$ is not active. Let $\tilde w:=w-\delta\mathbbm{1}_U$. Then $\tilde w\geq v$ and $(\bar G\tilde w)(s,a)=(\bar Gw)(s,a)-\delta\sum_{s'\in U}p(s,a,s')$, so, using \eqref{eq:hit_closed_U} in the second line,
    \begin{align*}
        s\notin U:&\quad (\bar G\tilde w)(s,a)\leq(\bar Gw)(s,a)\leq w(s)=\tilde w(s),\\
        s\in U,\ a\text{ active}:&\quad (\bar G\tilde w)(s,a)=(\bar Gw)(s,a)-\delta=w(s)-\delta=\tilde w(s),\\
        s\in U,\ a\text{ not active}:&\quad (\bar G\tilde w)(s,a)\leq(\bar Gw)(s,a)\leq w(s)-\delta=\tilde w(s).
    \end{align*}
    Hence $\tilde w\geq V_{\bar G\tilde w}$ is a super-solution above $v$ with $\tilde w<w$ on $U$, which contradicts the minimality of $w$.

    \emph{Step 3:} Let $d(s)$ be the number of edges of a shortest path from $s$ to $C_0$ in $\mathcal G$, so that $d=0$ on $C_0$ and $1\leq d(s)\leq|\cS|-1$ otherwise. For $s\notin C_0$, pick an edge $s\to s'$ with $d(s')=d(s)-1$, and let $\widehat\pi(s)$ be an action active at $s$ with $p(s,\widehat\pi(s),s')>0$. On $C_0$, choose $\widehat\pi(s)\in\arg\max_a(\bar Gw)(s,a)$. Then $\widehat\pi\in\Pi_{\mathrm{SD}}$ and, with $p_{\min}:=\min\{p(s,a,s'):p(s,a,s')>0\}$,
    \begin{align*}
        \bP_s^{\widehat\pi}\br{\tau_{C_0}\leq|\cS|}\geq p_{\min}^{|\cS|}=:\beta>0,\qquad s\in\cS.
    \end{align*}
    By the Markov property, $\bP_s^{\widehat\pi}(\tau_{C_0}>m|\cS|)\leq(1-\beta)^m$ for $m\geq1$. Hence $\bP_s^{\widehat\pi}(\tau_{C_0}<\infty)=1$.
    
    To apply \eqref{def:finite_hitting_time}, we show that $\widehat\pi$ is optimal from every state. Outside $C_0$, its actions are active, so $F_{\widehat\pi}w=w$. On $C_0$, monotonicity and $w=v$ give
    \begin{align*}
        w\geq F_{\widehat\pi}w=V_{\bar Gw}\geq V_{\bar Gv}\geq v-\rho\bo=w-\rho\bo.
    \end{align*}
    Thus $0\leq w-F_{\widehat\pi}w\leq\rho\bo$ on all of $\cS$. Telescoping this inequality along trajectories under $\widehat\pi$, dividing by the horizon, and using boundedness of $w$ yields
    \begin{align*}
        0\leq J\ust-J(\widehat\pi,s)\leq\rho<\Delta_{\rm pol},\qquad s\in\cS.
    \end{align*}
    By \eqref{def:policy_gap}, $\widehat\pi$ is optimal from every state. Consequently, \eqref{def:finite_hitting_time} gives
    \begin{align}
        \bE_s^{\widehat\pi}[\tau_{C_0}]\leq\mathcal T,\qquad s\in\cS.\label{eq:hit_envelope_time}
    \end{align}

    \emph{Step 4:} Write $\tau:=\tau_{C_0}$, $\bE_s:=\bE_s^{\widehat\pi}$, and $g(s):=\ol r(s,\widehat\pi(s))-J\ust$, so that $\abs{g}\leq1$ because $\ol r,J\ust\in[0,1]$. Since $\widehat\pi(s)$ is active at $s\notin C_0$, and $(\bar Gv)(s,a)-v(s)\leq\rho$ for all $(s,a)$,
    \begin{align*}
        w(s)=g(s)+(P^{\widehat\pi}w)(s),\quad s\notin C_0,\qquad
        v(s)\geq g(s)-\rho+(P^{\widehat\pi}v)(s),\quad s\in\cS.
    \end{align*}
    Fix $k\geq1$. For $t<k$, the event $\{t<\tau\}$ depends only on $S_0,\dots,S_t$, and $S_t\notin C_0$ on it. By the Markov property,
    \begin{align*}
        \bE_s\sqbr{\br{w(S_t)-w(S_{t+1})}\mathbbm{1}_{\{t<\tau\}}}&=\bE_s\sqbr{g(S_t)\mathbbm{1}_{\{t<\tau\}}},\\
        \bE_s\sqbr{\br{v(S_t)-v(S_{t+1})}\mathbbm{1}_{\{t<\tau\}}}&\geq\bE_s\sqbr{\br{g(S_t)-\rho}\mathbbm{1}_{\{t<\tau\}}}.
    \end{align*}
    Summing over $t=0,\dots,k-1$, the left-hand sides telescope:
    \begin{align*}
        w(s)=\bE_s\sqbr{\sum_{t=0}^{\tau\wedge k-1}g(S_t)+w(S_{\tau\wedge k})},\qquad
        v(s)\geq\bE_s\sqbr{\sum_{t=0}^{\tau\wedge k-1}\br{g(S_t)-\rho}+v(S_{\tau\wedge k})}.
    \end{align*}
    Let $k\to\infty$. The sums are bounded in absolute value by $(1+\rho)\tau$, which is integrable by \eqref{eq:hit_envelope_time}, $w$ and $v$ are bounded, and $\tau<\infty$ almost surely. By dominated convergence, and since $w(S_\tau)=v(S_\tau)$ because $S_\tau\in C_0$,
    \begin{align*}
        w(s)=\bE_s\sqbr{\sum_{t=0}^{\tau-1}g(S_t)+v(S_\tau)},\qquad
        v(s)\geq\bE_s\sqbr{\sum_{t=0}^{\tau-1}g(S_t)+v(S_\tau)}-\rho\bE_s[\tau].
    \end{align*}
    Subtracting and using \eqref{eq:hit_envelope_time} gives $w(s)-v(s)\leq\rho\bE_s[\tau]\leq\mathcal T\rho$. Together with $w\geq v$ from Step 1, this proves \eqref{eq:hit_envelope}.\qedhere
\end{proof}

%% file: appendix/mds_props.tex
\section{A Martingale Inequality in the Sup-Norm}
\label{sec:app_mds}

Condition~\eqref{eq:mds_const} holds on any finite-dimensional normed space, with a constant obtained from equivalence with a Euclidean norm. For the sup-norm, that yields $\nu = d$. The following lemma establishes that $\nu = O(\log(d))$, which is used in Theorem~\ref{thm:rvi_main}. It is the sup-norm form of the smooth-$\ell^p$ estimate used by \citet{chen2020finite}.

\begin{lemma}[Martingale constant in the sup-norm]\label{lem:mds_sup}
    Let $d \geq 2$, let $\cX=\bR^d$ be equipped with $\norm{\cdot}_\infty$, and let $\{U_i\}$ be a finite $\cX$-valued martingale difference sequence. Then, 
    \begin{align*}
        \bE\norm{\sum_{i}{U_i}}^2 \leq 2e\log(2d) \sum_i{\bE\norm{U_i}^2}.
    \end{align*}
\end{lemma}

\begin{proof}
    Let $p \geq 2$. The function $x \mapsto \frac12\norm{x}_p^2$ is convex and $(p-1)$-smooth with respect to $\norm{\cdot}_p$~\citep{beck2017first}, so the descent lemma applied along the partial sums, followed by conditioning and telescoping, gives the standard estimate
    \begin{align*}
        \bE\norm{\sum_i{U_i}}_p^2 \leq (p-1)\sum_i{\bE\norm{U_i}_p^2}.
    \end{align*}
    Combining it with $\norm{x}_\infty \leq \norm{x}_p \leq d^{1/p}\norm{x}_\infty$ yields
    \begin{align*}
        \bE\norm{\sum_i{U_i}}_\infty^2 \leq (p-1)\,d^{2/p}\sum_i{\bE\norm{U_i}_\infty^2}.
    \end{align*}
    Choose $p = \max\{2,\,2\log(d)\}$. If $d \geq e$ then $d^{2/p} = d^{1/\log(d)} = e$ and the prefactor is $(2\log(d) - 1)e \leq 2e\log(2d)$; if $d < e$, that is $d = 2$, the prefactor is $(2-1)\cdot 2 = 2 \leq 2e\log(2d)$. In either case $\bE\norm{\sum_i{U_i}}_\infty^2 \leq 2e \log(2d) \sum_i{\bE\norm{U_i}_\infty^2}$, which proves the claim.\qedhere
\end{proof}

%% file: appendix/stepsize.tex
\section{Stepsize Estimates}
\label{sec:app_aux}

The following elementary estimates are the only properties of the stepsize sequence \eqref{eq:stepsize} used in Section~\ref{sec:km}. Bounds \eqref{eq:A_n_poly} and \eqref{eq:A_n_harmonic} convert the accumulated stepsize into an explicit rate; bounds \eqref{eq:block_length} and \eqref{eq:ratio} are what keep the blocks of \eqref{def:blocks} from growing by more than a factor two on the shifted clock $n \mapsto n+K_0$.

\begin{lemma}[Stepsize sums and block growth]\label{lem:stepsize_sums}
    Let the stepsizes satisfy Assumption~\ref{assum:stepsize}. If $a \in (0,1)$, then for every $n \geq 0$,
    \begin{align}
        \sum_{k=1}^n\alpha_k \geq \frac{c}{1-a}\sqbr{(n+K_0+1)^{1-a} - (K_0+1)^{1-a}}, \label{eq:A_n_poly}
    \end{align}
    while if $a = 1$, then for every $n \geq 0$,
    \begin{align}
        \sum_{k=1}^n\alpha_k \geq c\log\br{\frac{n+K_0+1}{K_0+1}}. \label{eq:A_n_harmonic}
    \end{align}
    If in addition the first inequality in \eqref{eq:stepsize_cond} holds, then for every integer $m \geq 0$,
    \begin{align}
        \sum_{k=m+1}^{2m+K_0}{\alpha_k} \geq \frac{cK_0^{1-a}}{2} \geq L. \label{eq:block_length}
    \end{align}
    Finally, for all integers $m \leq n \leq 2m+K_0$,
    \begin{align}
        \alpha_{m+1} \leq 2^a\alpha_{n+1} \leq 2\alpha_{n+1}. \label{eq:ratio}
    \end{align}
\end{lemma}

\begin{proof}
    Since $t \mapsto t^{-a}$ is decreasing, $c(k+K_0)^{-a} \geq \int_{k+K_0}^{k+K_0+1}{ct^{-a}\,dt}$, and summing over $k = 1,\ldots,n$ gives
    \begin{align*}
        \sum_{k=1}^n\alpha_k \geq \int_{K_0+1}^{n+K_0+1}{ct^{-a}\,dt},
    \end{align*}
    which is \eqref{eq:A_n_poly} for $a<1$ and \eqref{eq:A_n_harmonic} for $a=1$.

    For \eqref{eq:block_length}, the indices $k \in \{m+1,\ldots,2m+K_0\}$ satisfy $k+K_0 \leq 2(m+K_0)$, so each of these $m+K_0$ stepsizes is at least $c\,(2(m+K_0))^{-a}$ and
    \begin{align*}
        \sum_{k=m+1}^{2m+K_0}{\alpha_k} \geq \frac{c\,(m+K_0)}{(2(m+K_0))^a} = \frac{c\,(m+K_0)^{1-a}}{2^a} \geq \frac{c\,K_0^{1-a}}{2} \geq L,
    \end{align*}
    using $m+K_0 \geq K_0$ together with $1-a \geq 0$, and $2^a \leq 2$.

    For \eqref{eq:ratio}, the hypothesis $n \leq 2m+K_0$ gives $n+1+K_0 \leq 2m+2K_0+1 < 2(m+1+K_0)$, hence
    \begin{align*}
        \frac{\alpha_{m+1}}{\alpha_{n+1}} = \br{\frac{n+1+K_0}{m+1+K_0}}^{a} \leq 2^a \leq 2. \quad\qedhere
    \end{align*}
\end{proof}